\documentclass[a4paper,fleqn]{cas-sc}

\usepackage[authoryear]{natbib}
\usepackage{graphicx} 
\usepackage{algorithm}  
\usepackage{algpseudocode}
\usepackage{color}
\usepackage{setspace}
\usepackage{subfiles}
\usepackage{caption}
\usepackage{subcaption}
\usepackage{bm}
\usepackage{amsmath}
\usepackage{amssymb}
\usepackage{amsfonts}
\usepackage{amstext}
\usepackage{amsthm}
\usepackage{soul}
\usepackage{hyperref}
\usepackage{ulem} 
\usepackage{float}

\def\tsc#1{\csdef{#1}{\textsc{\lowercase{#1}}\xspace}}
\tsc{WGM}
\tsc{QE}
\tsc{EP}
\tsc{PMS}
\tsc{BEC}
\tsc{DE}

\usepackage{lineno}

\begin{document}
\let\WriteBookmarks\relax
\def\floatpagepagefraction{1}
\def\textpagefraction{.001}
\shorttitle{Prediction of Effective Elastic Moduli using GNNs}
\shortauthors{Chung et al.}

\title [mode = title]{Prediction of Effective Elastic Moduli of Rocks using Graph Neural Networks}

\author[1]{Jaehong Chung}[type=editor,
auid=000,bioid=1,orcid=https://orcid.org/0000-0003-2960-4601]
\author[2]{Rasool Ahmad}[orcid=https://orcid.org/0000-0002-4154-6902]
\author[3]{WaiChing Sun}[orcid=https://orcid.org/0000-0002-3078-5086]
\author[2]{Wei Cai}[orcid=https://orcid.org/0000-0001-5919-8734]
\author[1, 4]{Tapan Mukerji}[type=editor, auid=000,bioid=1,orcid=https://orcid.org/0000-0003-1711-1850]

\address[1]{Department of Geophysics, Stanford University, Stanford, CA, USA.}
\address[2]{Department of Mechanical Engineering, Micro and Nano Mechanics Group, Stanford University, Stanford, CA, USA}
\address[3]{Department of Civil Engineering and Engineering Mechanics, Columbia University, New York, NY, USA}
\address[4]{Department of Energy Science and Engineering, Stanford University, Stanford, CA, USA}

\begin{abstract}
This study presents a Graph Neural Networks (GNNs)-based approach for predicting the effective elastic moduli of rocks from their digital CT-scan images. We use the Mapper algorithm to transform 3D digital rock images into graph datasets, encapsulating essential geometrical information. These graphs, after training, prove effective in predicting elastic moduli. Our GNN model shows robust predictive capabilities across various graph sizes derived from various subcube dimensions. Not only does it perform well on the test dataset, but it also maintains high prediction accuracy for unseen rocks and unexplored subcube sizes. Comparative analysis with Convolutional Neural Networks (CNNs) reveals the superior performance of GNNs in predicting unseen rock properties. Moreover, the graph representation of microstructures significantly reduces GPU memory requirements (compared to the grid representation for CNNs), enabling greater flexibility in the batch size selection. This work demonstrates the potential of GNN models in enhancing the prediction accuracy of rock properties and boosting the efficiency of digital rock analysis. 
\end{abstract}


\maketitle 

\printcredits

\doublespacing

\section{Introduction}
\label{intro}
Understanding and predicting rock mechanical properties, such as elastic moduli, plays a crucial role across a range of geoscience and engineering fields including energy resources engineering \citep{sone2013mechanical, madhubabu2016prediction}, geotechnical engineering \citep{zhang2021effect}, and seismology \citep{byerlee1968stick}. These properties govern how rocks react to in-situ stresses, shaping the macroscopic behaviors of geological structures. Accurate characterization of these properties is therefore instrumental in predicting phenomena such as seismic wave propagation, reservoir behavior, and slope stability. Fundamentally, these macroscopic behaviors originate from the intricate features at the microscopic scale. Specifically, the effective elastic moduli of rocks are influenced by three main factors: (1) the composition of pores and minerals, (2) the properties of these constituents, and (3) the geometric details of the rock's microstructure \citep{mavko2020rock}. While the first two factors can be relatively straightforwardly measured and characterized, capturing the complexity of the geometric details often presents a substantial challenge. Historically, traditional micromechanics homogenizations and rock physics relied on empirical relationships or theoretical models based on idealized microstructures to estimate rock properties \citep{li2008introduction, han1986effects, mindlin1949compliance, hill1952elastic}. While useful, these models often require overly idealized geometric assumptions such as the ellipsoidal inclusions \citep{hill1965self, mori1973average,berryman1980long,mura2013micromechanics,norris1985effective}, as well as approximation of multiple inclusion interactions which can potentially compromise their predictive power \citep{andra2013digitala, keehm2003computational}.

The advent of modern imaging techniques has enabled the rapid development of Digital Rock Physics (DRP), which offers a more reliable way of predicting rock properties. In contrast to traditional methods, DRP makes use of 3D rock images obtained through advanced imaging technologies such as micro-CT scanners, thus eliminating the need for idealized geometric assumptions \citep{andra2013digitala}. This approach embraces the complex and heterogeneous nature of geological materials, enabling accurate rock property predictions through pore-scale numerical simulations \citep{andra2013digitalb}. Consequently, the integration of imaging and computational techniques in DRP holds promise for a more detailed understanding of geological materials, revealing relationships between rock properties and pore-scale fundamental processes.

Despite these advancements, the computational burden associated with simulating rock properties across diverse geometries remains a significant hurdle. Physical simulations involving solutions of partial differential equations on complex geometries require substantial computational resources and time to generate satisfactory results. Deep learning, with its capability to harness geometric information, has emerged as a promising solution to this issue \citep{voulodimos2018deep}. Deep learning models that can extract geometrical features and insights from data have the potential to predict rock properties more efficiently and accurately. Therefore, combining deep learning with DRP could enhance rock property prediction accuracy and reduce computational costs, serving as a tool to unearth the non-linear geometrical effects on rock properties.

Previous works have demonstrated the potential of machine learning and deep learning techniques in predicting rock properties based on their microstructures, including elastic constants \citep{ahmad2023homogenizing}, permeability \citep{wu2018seeing, tembely2020deep, hong2020rapid, liu2023hierarchical}, and effective diffusivity \citep{wu2019predicting}. However, these techniques have their limitations. For example, Convolutional Neural Networks (CNNs), the main architecture in these approaches, are typically constrained to a fixed voxel size. This implies that any change in voxel size necessitates a new image dataset and labels for property prediction training. Moreover, CNNs can be computationally demanding, as they consider all pixels in the images, resulting in a large number of learnable parameters and memory. This requirement for extensive Graphics Processing Unit (GPU) memory not only demands resources but also affects training accuracy by constraining the batch size, a critical parameter for model optimization \citep{smith2017don, he2019control, kandel2020effect}. To address some of these limitations, PointNet architectures were introduced, which represent the microstructures with a point cloud to predict the permeability of porous media \citep{kashefi2021point}. Despite its more economic use of GPU memory compared to CNNs, PointNet also has its limitations. The main issue is that PointNet requires the same number of input points for all samples \citep{qi2017pointnet}, which imposes constraints when selecting an identical number of representative points for diverse rock microstructures. 
A recent study by \cite{alzahrani2023pore} combined CNN and Graph Neural Networks (GNN) to predict the permeability of digital rocks. Yet, their approach still relies on CNN to extract a fixed set of feature vectors, not taking advantage of the full potential of GNNs, which we highlight in this research.

In response to these challenges, this study proposes a novel approach that employs a GNN model for predicting the effective elastic moduli of digital rocks. The GNN offers several potential advantages for digital rock applications. Firstly, GNN can handle non-Euclidean data \citep{bronstein2017geometric}, potentially dealing with irregularities in the data more effectively than CNNs, which are primarily designed for grid-like structures. Secondly, unlike CNNs, GNNs represent microstructures efficiently through graphs rather than requiring all voxels as inputs. This sparse representation implies that GNNs can focus on key geometrical features within the materials such as solid connectivities and pore networks, rather than processing every voxel. For instance, a graph constructed from a $100^3$ voxel digital cube typically comprises of a few hundred nodes and edges, and hence requires significantly less memory---less than two orders of magnitude---compared to the memory needed for the entire $100^3$ voxel digital rock image. This approach also reduces GPU memory requirements for machine learning training, as it avoids storing and processing extensive voxel data. Thirdly, GNN can process graph inputs of different sizes, allowing one GNN model to be trained on and process diverse rock microstructures. 

This work's contributions are threefold: (1) proposing a novel framework to convert 3D digital rock images into graph data while preserving topological features using Mapper algorithms, (2) verifying the informativeness of the topological features of these graphs for mechanical property prediction, and (3) demonstrating the capability of GNNs in predicting the effective bulk and shear modulus of digital rocks, exhibiting computational efficiency and superior predictive accuracy compared to conventional CNN models.

The remainder of the paper is organized as follows: Section \ref{section:Data} provides detailed information about the digital rock dataset used in this study. Section \ref{section:Methods} describes our approach, including the Mapper algorithm and GNN architectures. Section \ref{section:Results} presents and discusses the results. Section \ref{section:Conclusion} concludes the study.

\section{Digital Rock Dataset and Effective Elastic Moduli Computation}
\label{section:Data}

\begin{figure}
    \centering
    \begin{subfigure}{0.49\textwidth}
        \includegraphics[width=\linewidth]{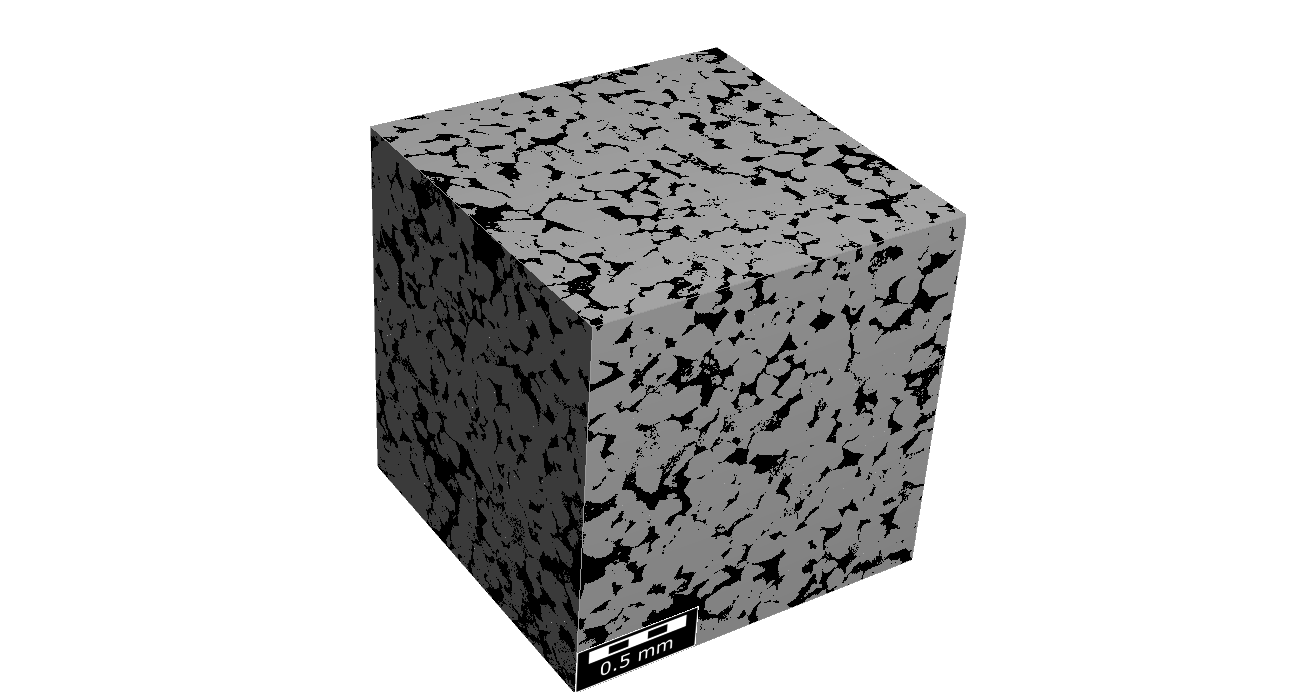}
        \caption{Berea sandstone 1 (B1)}
        \label{fig:B1}
    \end{subfigure}
    \begin{subfigure}{0.49\textwidth}
        \includegraphics[width=\linewidth]{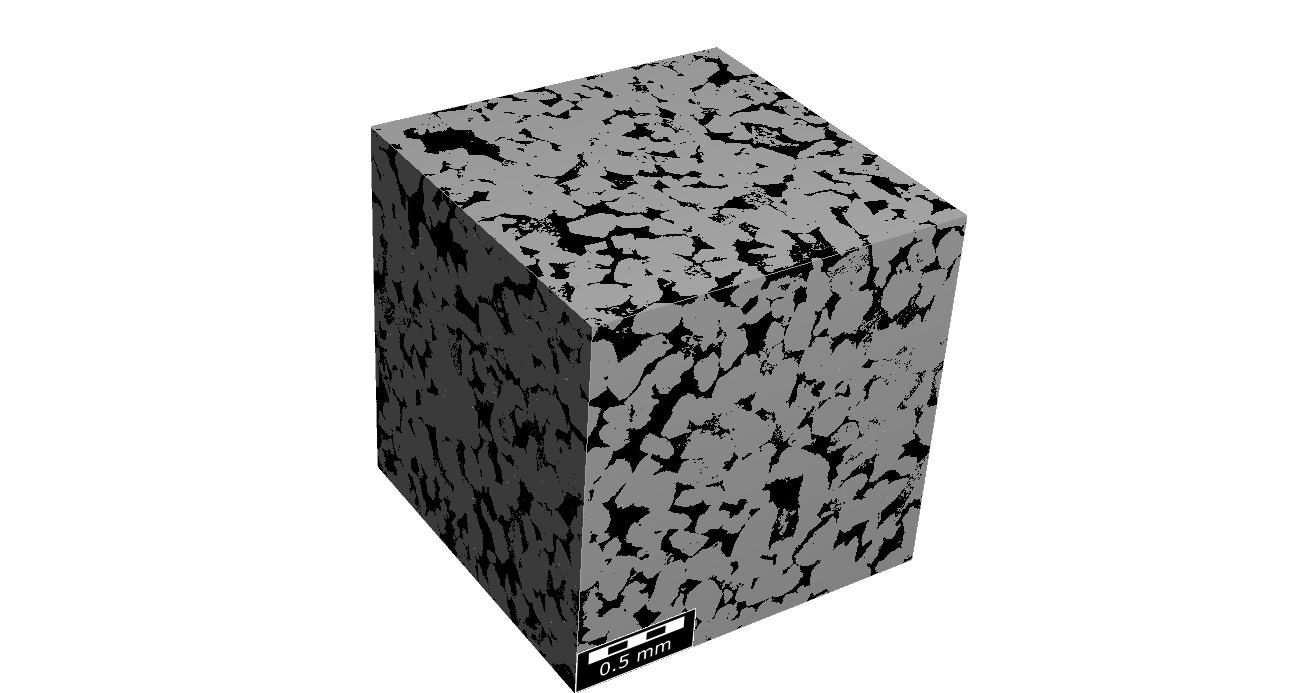}
        \caption{Berea sandstone 2 (B2)}
        \label{fig:B2}
    \end{subfigure}
    
    \begin{subfigure}{0.49\textwidth}
        \includegraphics[width=\linewidth]{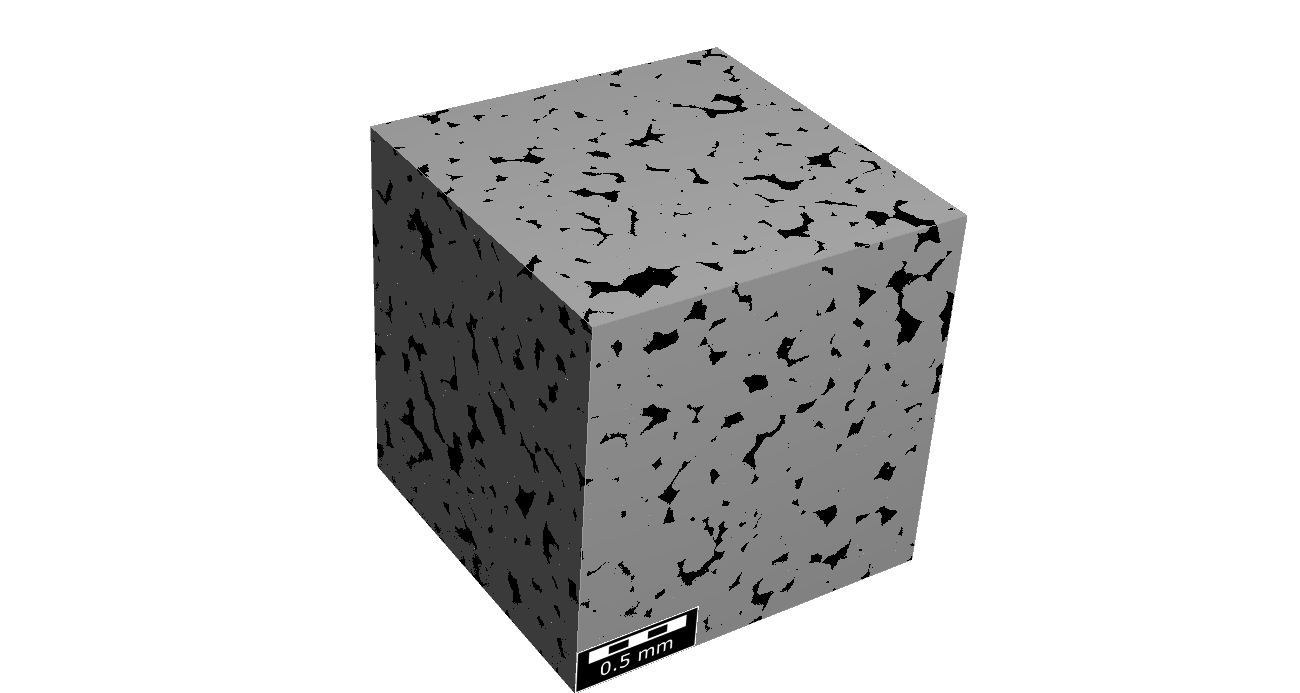}
        \caption{Fontainebleau sandstone 1 (FB1)}
        \label{fig:FB1}
    \end{subfigure}
    \begin{subfigure}{0.49\textwidth}
        \includegraphics[width=\linewidth]{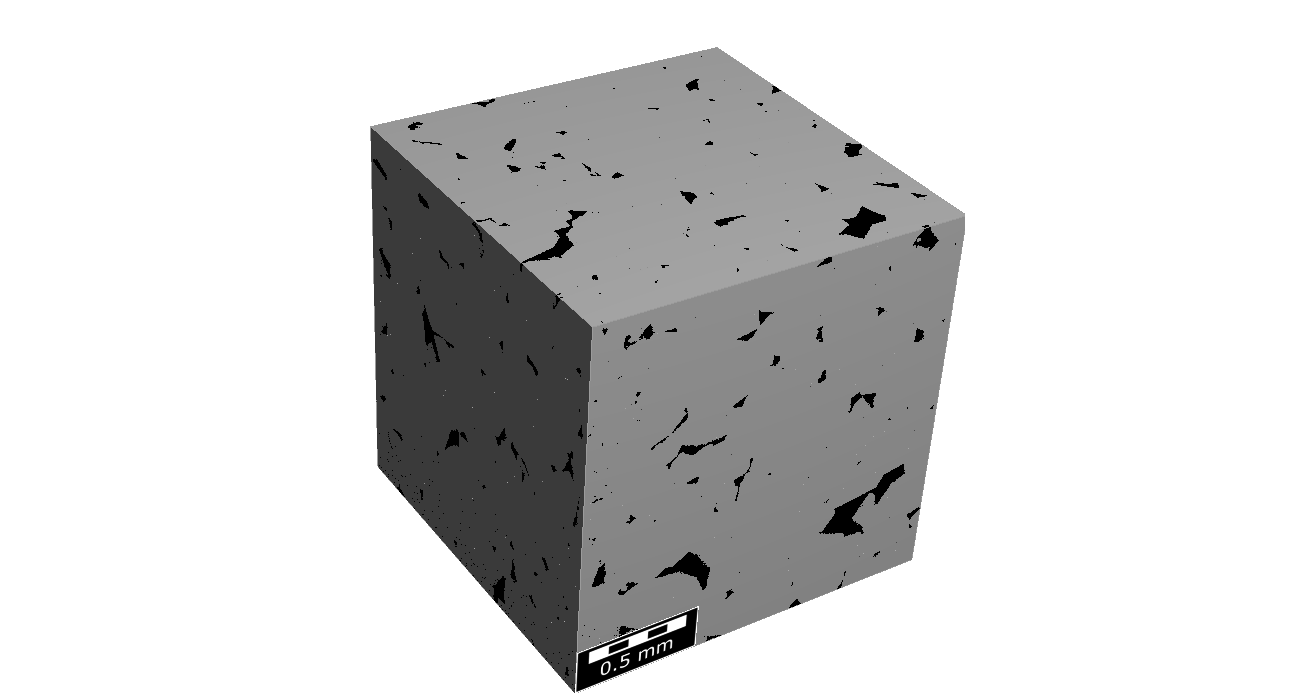}
        \caption{Fontainebleau sandstone 2 (FB2)}
        \label{fig:FB2}
    \end{subfigure}
    
    \begin{subfigure}{0.49\textwidth}
        \includegraphics[width=\linewidth]{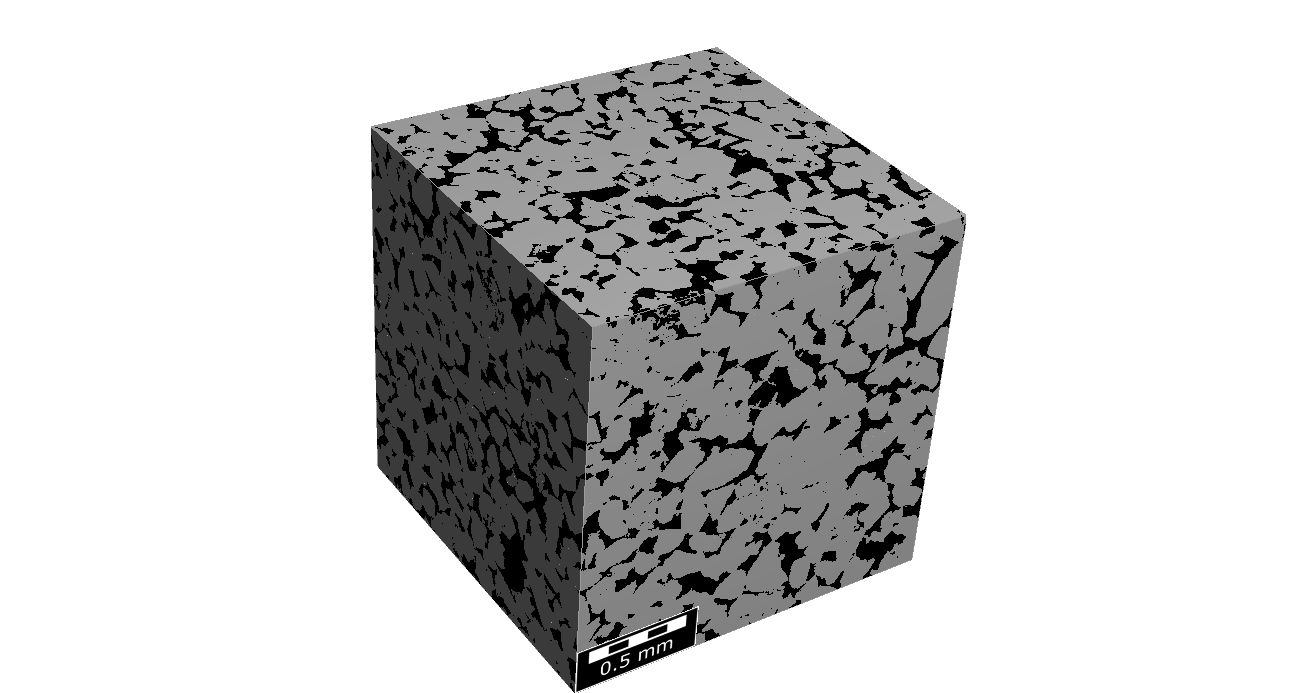}
        \caption{Castlegate sandstone (CG)}
        \label{fig:CG}
    \end{subfigure}
    
    \caption{Five digital rock images used in the study. Gray represents grain and black represents pore.}
    \label{fig:rock_samples}
\end{figure}

The rock data utilized in this study consists of five samples, which include two Berea sandstones (B1, B2), two Fontainebleau sandstones (FB1, FB2), and one Castlegate sandstone (CG). Each sample consists of $900^3$ voxels at a resolution of roughly $2\mu m$ as shown in Figure \ref{fig:rock_samples}. X-ray diffraction (XRD) analysis of the samples revealed a predominant composition of quartz mineral, with the remaining volume being pore space \citep{saxena2019rock}. For simplicity, we employed a binary phase model, treating all solid minerals as quartz and the other phase as air-filled pore space. We follow the multiscale framework to compute the effective mechanical properties of geomaterials from microscale features \citep{andra2013digitalb}. The elastic moduli data used for training and validation in this study were computed from our previous studies. In this section, we briefly introduce how to compute the mechanical properties. For more details, please see \cite{ahmad2023computation}. 

The computation of elastic moduli is conducted in two steps. Initially, a sub-image volume, denoted as $\Omega$, is extracted from the larger image. Subsequently, the elasticity partial differential equations (PDEs) are solved within the domain $\Omega$ under periodic boundary conditions (PBC). The stress and strain fields derived from this process are used to compute the average stress and strain values in $\Omega$, thereby obtaining its effective elastic constants. A brief description of the steps of this approach are given below.

The local elastic response within the domain $\Omega$ at a point $\bm{x}$ is represented by a fourth-order position-dependent isotropic elastic stiffness tensor $\mathbb{C}_{ijkl}(\bm{x})$. For example, the local elastic stiffness corresponds to that of quartz or air depending on the classification of the voxel at $\bm{x}$.
The elastic stiffness tensor relates the local stress $\sigma_{ij}(\bm{x})$ with the local elastic strain $\varepsilon_{ij}(\bm{x})$
through Hooke' law:
\begin{equation}
    \sigma_{ij}(\bm{x}) = \mathbb{C}_{ijkl}(\bm{x}) \, \varepsilon_{kl}(\bm{x}),
\end{equation}
where we work with Einstein's notation and the convention of summation over repeated indices.

The elastic stiffness tensor $\mathbb{C}_{ijkl}(\bm{x})$ of an isotropic medium is expressed in terms of the position dependent bulk $K(\bm{x})$ and shear  $\mu(\bm{x})$ moduli, i.e.
\begin{equation}
\mathbb{C}_{ijkl}(\bm{x}) = 
\left(K(\bm{x}) - \frac{2}{3}\mu(\bm{x}) \right) \delta_{ij}\delta_{kl} + \mu(\bm{x}) \left(\delta_{ik}\delta_{jl} + \delta_{il}\delta_{jk}\right),
\end{equation}
where $\delta_{ij}$ is the Kronecker delta symbol. Due to the periodic boundary condition, the elastic strain field $\varepsilon_{ij}(\bm{x})$ can be decomposed as the sum of an average macroscopic (i.e. homogeneous) part $E_{ij}$ and a locally fluctuating (i.e. heterogeneous) part $\varepsilon_{ij}^{*}(\bm{x})$
\begin{equation}
\varepsilon_{ij}(\bm{x}) = E_{ij} + \varepsilon_{ij}^{*}(\bm{x}),
\end{equation}
where by definition
\begin{equation}
\int_{\Omega } \varepsilon_{ij}^{*}(\bm{x}) \, d\bm{x} = 0.
\end{equation}
The fluctuation in the strain field can be determined by requiring the following two conditions to be satisfied:
\begin{equation}
\begin{aligned}
\varepsilon_{ij}^{*}(\bm{x}) &= \frac{1}{2} [u_{i,j}^{*}(\bm{x}) + u_{j,i}^{*}(\bm{x})] \quad \forall \bm{x} \in \Omega  \qquad \text{(compatibility condition)}, \\
\sigma_{i,j}(\bm{x}) &= 0 \quad \forall \bm{x} \in \Omega \qquad \text{(equilibrium condition)},
\end{aligned}
\end{equation}
where $u_i^*(\bm{x})$ is a displacement field periodic over $\Omega$.

The equilibrium stress and strain fields are obtained by numerically solving the above equations using a fast Fourier transform (FFT) based elasticity solver as implemented in the ElastoDict module of GeoDict software \citep{geodict2023}. The homogenized elastic constants in the domain $\Omega $ is computed by taking the average of the stress and strain fields as
\begin{equation}
\begin{aligned}
\langle\sigma\rangle_{ij} &= \frac{1}{|\Omega |}\int_{\Omega } \sigma_{ij}(\bm{x}) \, d\bm{x}, \\
\langle\varepsilon\rangle_{ij} &= \frac{1}{|\Omega |}\int_{\Omega } [E_{ij} + \varepsilon_{ij}^{*}(x)] \, d\bm{x} = E_{ij}.
\end{aligned}
\end{equation}

The homogenized elastic stiffness of the subimage, $\Omega $, satisfies the following condition:
\begin{equation}
\langle\sigma\rangle_{ij} = \mathbb{C}_{ijkl}^ {\rm homo}\langle\varepsilon\rangle_{ij}.
\end{equation}
The homogenized stress-strain relation can then be expressed in the Voigt form as $\underline{\langle\sigma\rangle} = \underline{\underline{C}}^ {\rm homo} \underline{\langle\varepsilon\rangle}$, where $\underline{\langle\sigma\rangle}$ and $\underline{\langle\varepsilon\rangle}$ are $6\times 1$ column matrices and $\underline{\underline{C}}^ {\rm homo}$ is a $6\times 6$ matrix. To compute the homogenized stiffness matrix $\underline{\underline{C}}^ {\rm homo}$, we impose six independent macroscopic strain fields, resulting in six pairs of $\underline{\langle\sigma\rangle}$ and $\underline{\langle\varepsilon\rangle}$ vectors upon numerically solving the elasticity PDEs. Arranging these vectors into 6 $\times$ 6 stress $\underline{\underline{\Sigma}}$ and strain $\underline{\underline{\mathcal{E}}}$ matrices, we obtain the stress-strain relation, $\underline{\underline{\Sigma}} = \underline{\underline{C}}^ {\rm homo} \underline{\underline{\mathcal{E}}}$, from which the homogenized elastic stiffness tensor can be obtained as
\begin{equation}\underline{\underline{C}}^ {\rm homo} = \underline{\underline{\Sigma}}~\underline{\underline{\mathcal{E}}}^{-1}.
\end{equation}
Subsequently, the effective bulk and shear moduli are computed from $\underline{\underline{C}}^ {\rm homo}$ using the Voigt average \citep{hale2018evaluating, ahmad2023computation} as follows.
\begin{equation}
K^ {\rm homo} = \frac{(C^{\rm homo}_{11}+C^ {\rm homo}_{22}+C^ {\rm homo}_{33}) + 2(C^ {\rm homo}_{12}+C^ {\rm homo}_{13}+C^ {\rm homo}_{23})}{9}
\end{equation}
\begin{equation}
\mu^ {\rm homo} = \frac{(C^ {\rm homo}_{11}+C^ {\rm homo}_{22}+C^ {\rm homo}_{33}) - (C^ {\rm homo}_{12}+C^ {\rm homo}_{23}+C^ {\rm homo}_{13}) + 3(C^ {\rm homo}_{44}+C^ {\rm homo}_{55}+C^ {\rm homo}_{66})}{9}
\end{equation}

where the $C^ {\rm homo}_{ij}$ represents the components of matrix $\underline{\underline{C}}^ {\rm homo}$.


\section{Methods for Machine Learning}
\label{section:Methods}
\begin{figure}
\centering
\includegraphics[width=1.0\textwidth]{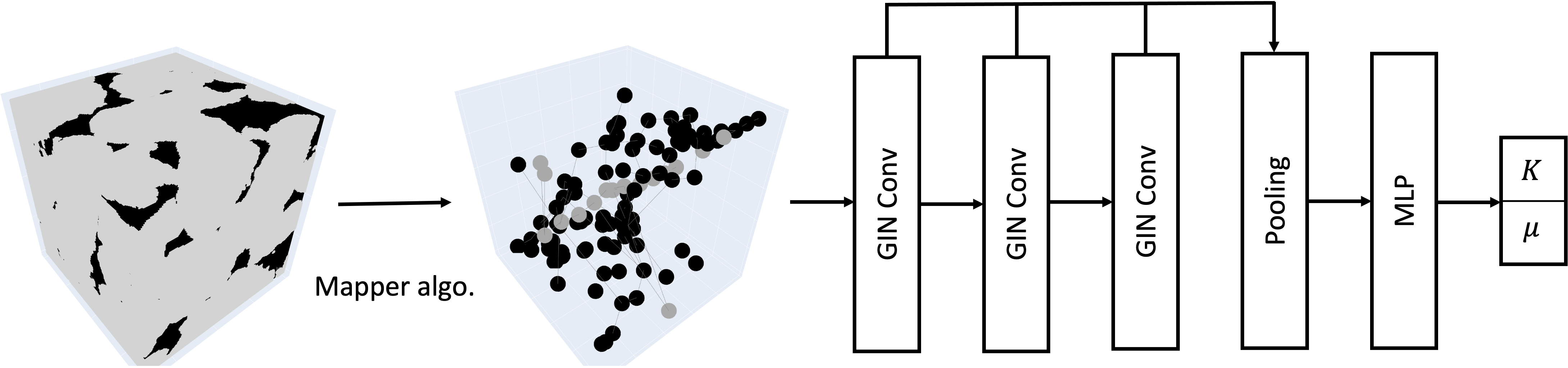}
\caption{Workflow: Conversion of a digital rock into a graph via the Mapper algorithm, followed by the utilization of GNN for elastic moduli prediction}
\label{fig:GNN_arch}
\end{figure}

There are two major parts of the workflow for training the machine learning model (Figure \ref{fig:GNN_arch}). In the first part, we construct a graph representation from the 3D segmented binary image of the rock. This is done using the Mapper algorithm described in section \ref{section:Mapper_algo}. The graph, which is a sparse representation of the pore space geometry and connectivity, is the input to the  supervised machine learning model, which is trained in the second part of the workflow. We compare the performances of two alternative approaches of training. In the first approach, features extracted from the graph (graph metrics) are used as predictors for a supervised random forest model. In the second approach, we directly feed the graph to a graph neural network (GNN) which is trained in a supervised learning mode to predict elastic moduli. The architecture for the GNN is described in section \ref{section:GNN_architecture}. 
As a benchmark, we also compare the GNN-based approaches against a CNN-based approach described in section \ref{section:CNN_architecture}.

\subsection{Mapper algorithm}
\label{section:Mapper_algo}
The Mapper algorithm is a topological data analysis tool, which transforms high-dimensional data into a simplified yet informative graph network structure \citep{singh2007topological}. By capturing both local and global geometric features, this algorithm can provide a comprehensive topological summary of complex data \citep{chazal2021introduction}. The capability of the algorithm has been validated for diverse complex data such as river delta patterns \citep{nesvold2019building}, cave morphology \citep{kanfar2023stochastic} and micro-bioscience image \citep{liao2019tmap}. In this study, we employed the Mapper algorithm to generate graph data from 3D digital rock images. Within the realm of digital rocks physics, this method looks particularly promising as it captures the multiresolution topology of the rock microstructure, an essential determinant of the rock's mechanical properties.

Data transformation into graph representations via the Mapper algorithm consists of several stages. Initially, the voxels of the CT image, represented in the 3D space $\bm{X}$, are projected into a lower-dimensional space $\bm{\chi}$ using a filter function $f: \bm{X} \rightarrow \bm{\chi}$. In this study, the filter function simply selects the $x$-component of the 3D voxel coordinates $\bm{X}(x, y, z)$, i.e. $\bm{\chi} = x$.
The purpose of defining the $\bm{\chi}$ space is to partition the original data into overlapping subsets.
%
This is accomplished by partitioning the lower-dimensional space $\bm{\chi}$ into overlapping subsets, ${u_i}$, $i=1, \cdots, N$, and $N$ is the total number of subsets.
Correspondingly, the original data set in 3D space $\bm{X}$ is also partitioned into overlapping subsets, ${U_i}$, $i=1, \cdots, N$.
The data in each of these subsets is then subjected to a clustering process, grouping together adjacent points (having the same label) within each subset $U_i$ into clusters $C_{ij}$, where $j$ denotes the individual clusters within the subset. In this study, the Depth-First Search (DFS) algorithm was utilized for identifying clusters \citep{tarjan1972depth,ryu2010numerical}. Upon cluster formation, each cluster $C_{ij}$ corresponds to a node in the graph. Edges are then established between nodes if the corresponding clusters share common points within the point cloud, i.e., an edge $(C_{ij}, C_{kl})$ exists if $C_{ij} \cap C_{kl} \neq \emptyset$. 

\begin{figure}[pos=h!]
\centering
\includegraphics[width=\textwidth]{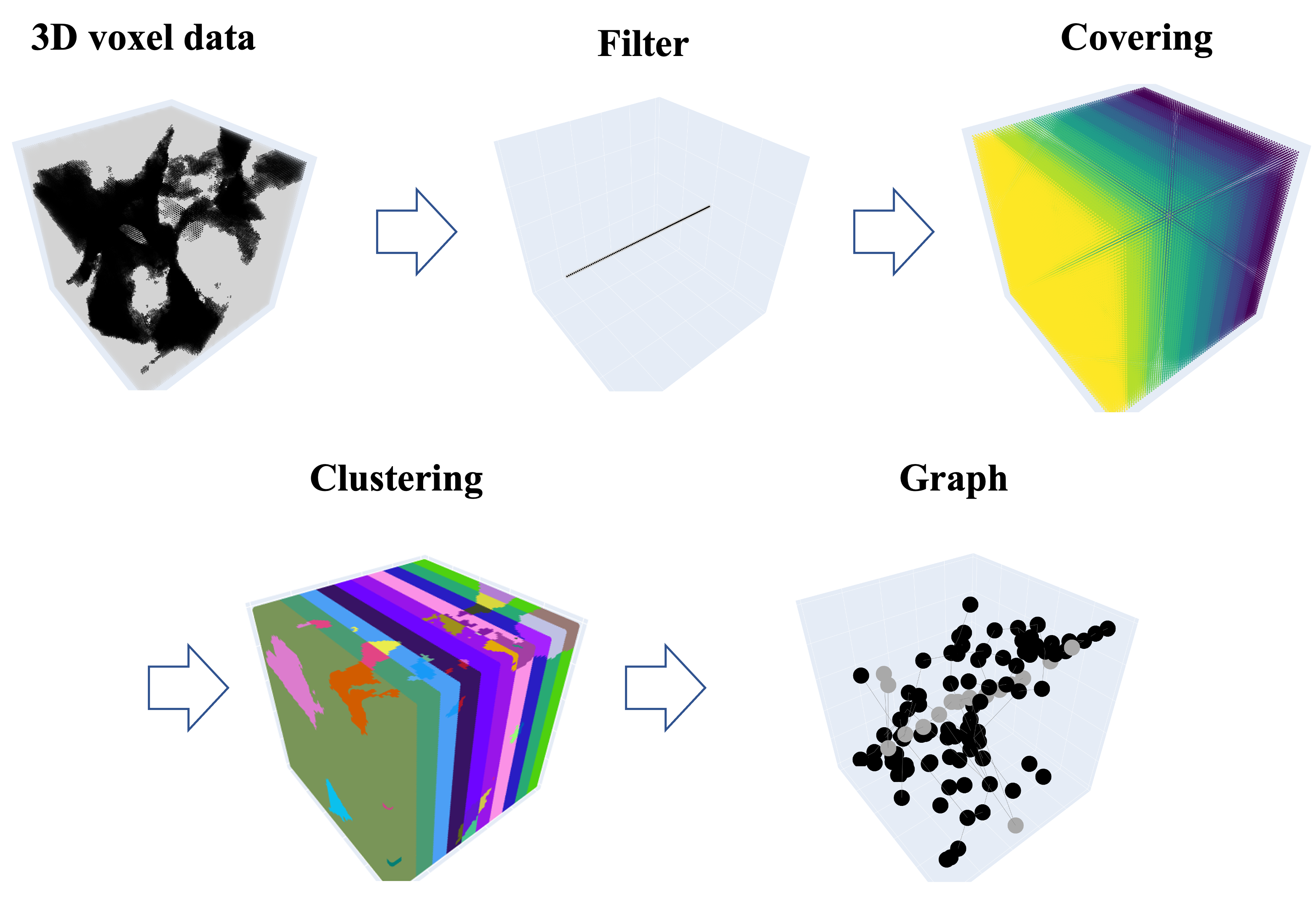}
\caption{The process of graph construction using the Mapper algorithm: (a) Original 3D voxel data where gray represents grain and black represents pore, (b) Projection on the $X$-axis using the filter function, (c) Covering procedure dividing the domain into 10 regions, (d) Clustering of pore and solid structures using DFS algorithm, (e) The resulting graph $G(V, E)$}
\label{fig:mapper}
\end{figure}

To accommodate the binary voxel data—grain and pore—the Mapper algorithm is run twice separately, and the graphs are concatenated with a node indicator to indicate a grain node or a pore node. The outcome is the graph data $G(V, E)$, where $V$ is the set of nodes representing either solid or pore structures, and $E$ is the set of undirected edges. It is noteworthy that this framework can potentially be extended to many-phase materials beyond the binary model by embedding material phase information into node features. After constructing the graphs, we extract topological features of graphs and embed this into node vector space. As listed in the Table \ref{tab:NodeFeatureVector}, each node feature vector is composed of 12 elements that represent a mixture of geometric, topological, and material properties, i.e., each feature vector $h_v \in \mathbb{R}^{12 \times 1}$. Detailed information on the computation of each topological feature is included in Appendix \ref{section:Appendix_graph_topo_metrics}. Figure \ref{fig:mapper} illustrates process of the graph data construction from digital rock using the Mapper algorithm.

\begin{table}
\centering
\caption{Node feature vector composition for the Graph Neural Network.}
\label{tab:NodeFeatureVector}
\begin{tabular}{ |p{3cm}||p{8cm}|p{2cm}|} 
 \hline
\textbf{Feature} & \textbf{Description} & \textbf{Dimensions} \\
 \hline 
 \hline
Center Coordinates & Spatial position of the node (x, y, z) & 3 \\
\hline
Cluster Dimensions & Size of the cluster (a, b, c) & 3 \\
\hline
Number of Points & Total points in the cluster & 1 \\
\hline
Node Degree & Edges connected to the node & 1 \\
\hline
Closeness Centrality & Inverse sum of shortest distances between the node and all others & 1 \\
\hline
Eigenvector Centrality & Measure of the node's influence based on its connections' quality and quantity & 1 \\
\hline
Pagerank & Importance of the node in the network based on its links and the significance of its neighboring nodes & 1 \\
\hline
Phase & Representation for phase (Solid = 1, Pore = 0, and vice versa) & 1 \\
\hline
\end{tabular} 
\end{table}

\subsection{Neural network architectures}
\subsubsection{Graph Neural Networks}
\label{section:GNN_architecture}

Graph Neural Networks (GNNs) are a specialized class of deep neural networks, designed to process graph-structured data. These networks use node neighbor aggregation and graph-level pooling mechanisms to derive node and graph-level features, respectively \citep{hamilton2020graph}. In a GNN, each node $u$ in the set of nodes $V$ is represented as a feature vector $h_u^{(k)}$, where $k$ denotes the iteration number. The feature vectors are updated through an iterative exchange of messages with neighboring nodes by neural network operations \citep{gilmer2017neural} such as follows.
\begin{equation}
\label{eqn:GNN}
h_v^{(k)} = \text{UPDATE}^{(k)} (h_v^{(k-1)}, \text{AGGREGATE}^{(k)}({h_u^{(k-1)}, \forall u \in \mathcal{N}(v)}))
\end{equation}
where $\mathcal{N}(v)$ represents the set of neighboring nodes of $v$. The AGGREGATE function uses the embedding of the nodes to generate the message from the aggregated information, and the UPDATE function combines previous embedding $h_v^{(k-1)}$ with the aggregated message to update the embedding of the node $v$. After that, graph-level feature representations can be obtained through the READOUT function that aggregates the set of node embedding from the last iteration $K$ \citep{zhang2018end, xu2018powerful}.
\begin{equation}
\label{eqn:GNN2}
h_G = \text{READOUT}(\{h_v^{(K)} \}\,|\, v \in G)
\end{equation}
In this study, we use the Graph Isomorphism Network (GIN), a specific variant of GNN, for learning the representations of graphs constructed from digital rock images. The GIN augments the capacity of GNNs by ensuring that isomorphic graph structures map to identical representations and non-isomorphic structures to distinct ones as powerful as the Weisfeiler Lehman graph isomorphism test \citep{xu2018powerful}. This is achieved through a unique update rule involving multi-layer perceptrons (MLPs) and a learnable parameter, $\epsilon^k$. Here, MLPs serve as function approximators in the network, mapping a set of input features to output features at each layer and thereby, enabling the learning of non-linear representations of the input graph. In addition the parameter $\epsilon^k$ modulates the relative importance of a node's own features and its neighbors' features during the aggregation process. The GIN update node representation is defined as follows:
\begin{equation}
\label{eqn:GIN}
h_v^{\text{GIN}(k)} = \text{MLP}^{(k)}((1 + \epsilon^k) \cdot h_v^{\text{GIN}(k-1)} + \Sigma_{u\in \mathcal{N}(v)} \, h_u^{\text{GIN}(k-1)})
\end{equation}
The graph-level feature is obtained by summing the node features at each iteration, which are then concatenated to yield the graph-level feature:
\begin{equation}
h_G^{\text{GIN}(k)}= \text{SUM}(\{h_v^{\text{GIN}(k)}\} \,|\, v\in G) \,| \, k = 0, 1, ..., K)
\end{equation}
\begin{equation}
h_G^{\text{GIN}}= \text{CONCAT}(\{{h_G^{\text{GIN}(k)}}\} \, | \, k = 0, 1, ..., K)
\end{equation}
This GIN operation ensures that isomorphic graphs have the same function output, a feature not guaranteed by other types of Graph Convolutional Networks (GCNs) \citep{xu2018powerful, zhang2019graph}. Hence, the GIN enhances the network's ability to capture and leverage graph structures during training. In our architecture, GIN layers are used to extract and learn from the graph-level features. Afterward, a pooling operation is used to condense the graph representation into a fixed-size vector, which is then processed by a Multi-Layer Perceptron (MLP) to output two scalar values, representing two effective elastic moduli of the respective rock sample. The GNN architecture is listed in the Table \ref{tab:Model_Comparison}.

\begin{table}
\centering
\caption{Comparison of the GNN and CNN models' structure and parameters.}
\label{tab:Model_Comparison}
\begin{tabular}{|c||c|c|} 
 \hline
 Contents & GNN & CNN \\ 
 \hline
 \hline
 Input layer & Node features (12 dim) & 3D image (1 channel) \\
 \hline
 Convolutional layers 1 & GIN Conv + ReLU (16 channel) & 3D Conv + ReLU (16 channel) \\
 \hline
 Convolutional layers 2 & GIN Conv + ReLU (16 channel) & 3D Conv + ReLU (16 channel) \\
 \hline
 Convolutional layers 3 & GIN Conv + ReLU (16 channel) & 3D Conv + ReLU (16 channel) \\
 \hline
 Pooling & Global add pooling & Max Pooling (2,2,2) \\
 \hline
 Fully connected layer 1 & Linear (48 dim input, 32 output) & Linear (calculated input size, 32 dim output) \\
 \hline
 Fully connected layer 2 & Linear (32 dim input, 32 dim output) & Linear (32 dim input, 32 dim output) \\
 \hline
 Fully connected layer 3 & Linear (32 dim input, 2 dim output) & Linear (32 dim input, 2 dim output) \\
 \hline
 Dropout & 0.3 & 0.3 \\
 \hline
 Activation function & ReLU & ReLU \\
 \hline
\end{tabular} 
\end{table}

\subsubsection{Convolutional Neural Networks}
\label{section:CNN_architecture}
\begin{figure}
\centering
\includegraphics[width=0.6\textwidth]{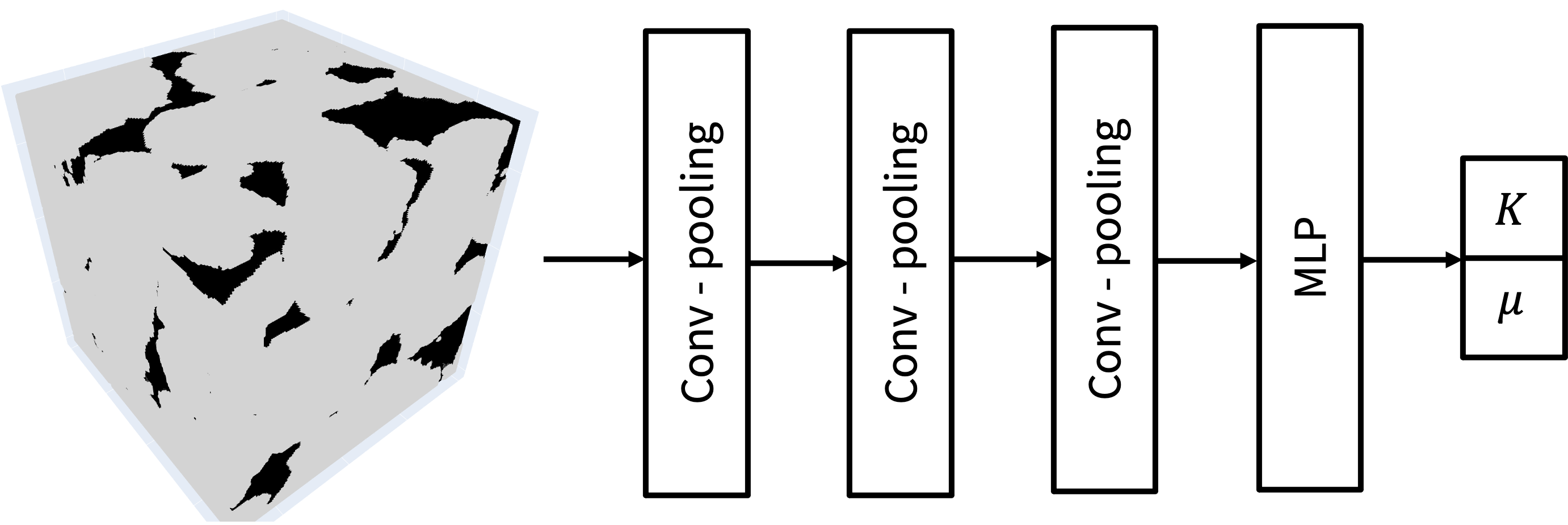}
\caption{CNN architecture for predicting effective elastic stiffness of rocks from 3D segmented images.}
\label{fig:CNN_arch}
\end{figure}

We also employed a CNN model to establish a benchmark against the GNN model. In this section, we briefly introduce the CNN architecture we designed to predict the effective elastic moduli of digital rocks. Specifically, our model consists of three convolutional layers, each equipped with 32 filters. We adopt a kernel size of (3,3,3) and a stride size of 2. Following the convolutional layers, the encoded image features are further processed through a decoder, which mirrors the structure used in the GNN architecture, except for the last convolutional layer due to the inherent data structure difference between GNN and CNN. For the training process, we utilize the Mean Squared Error (MSE) as the loss function and employ the Adam optimizer, again identical to the choices made for the GNN model. The CNN architecture used in this study is depicted in Figure \ref{fig:CNN_arch}. This design was chosen to closely mirror the GNN model structure, facilitating a fair comparison of the two techniques as much as possible.

\section{Results and Discussions}
\label{section:Results}
\subsection{Verification of Graph's Effectiveness in Property Prediction}
\label{section:random_forest}
\begin{figure}[pos=H]
    \centering
    \begin{subfigure}{0.45\textwidth}
    \centering
        \includegraphics[width=0.8\textwidth]{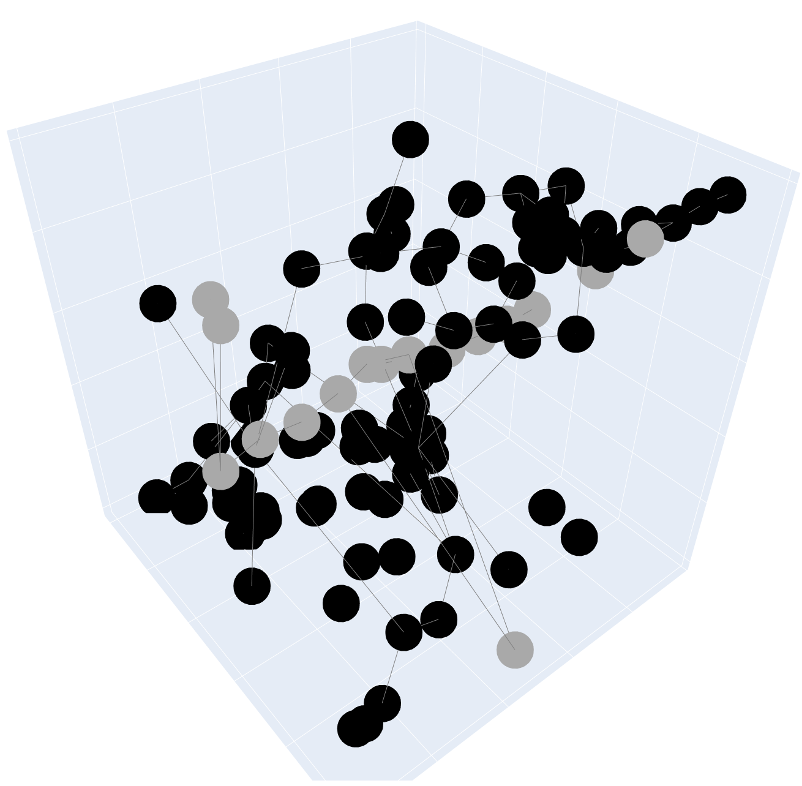}
        \label{fig:Figure1}
    \end{subfigure}
    \hfill
    \begin{subfigure}{0.45\textwidth}
    \centering
        \includegraphics[width=\textwidth]{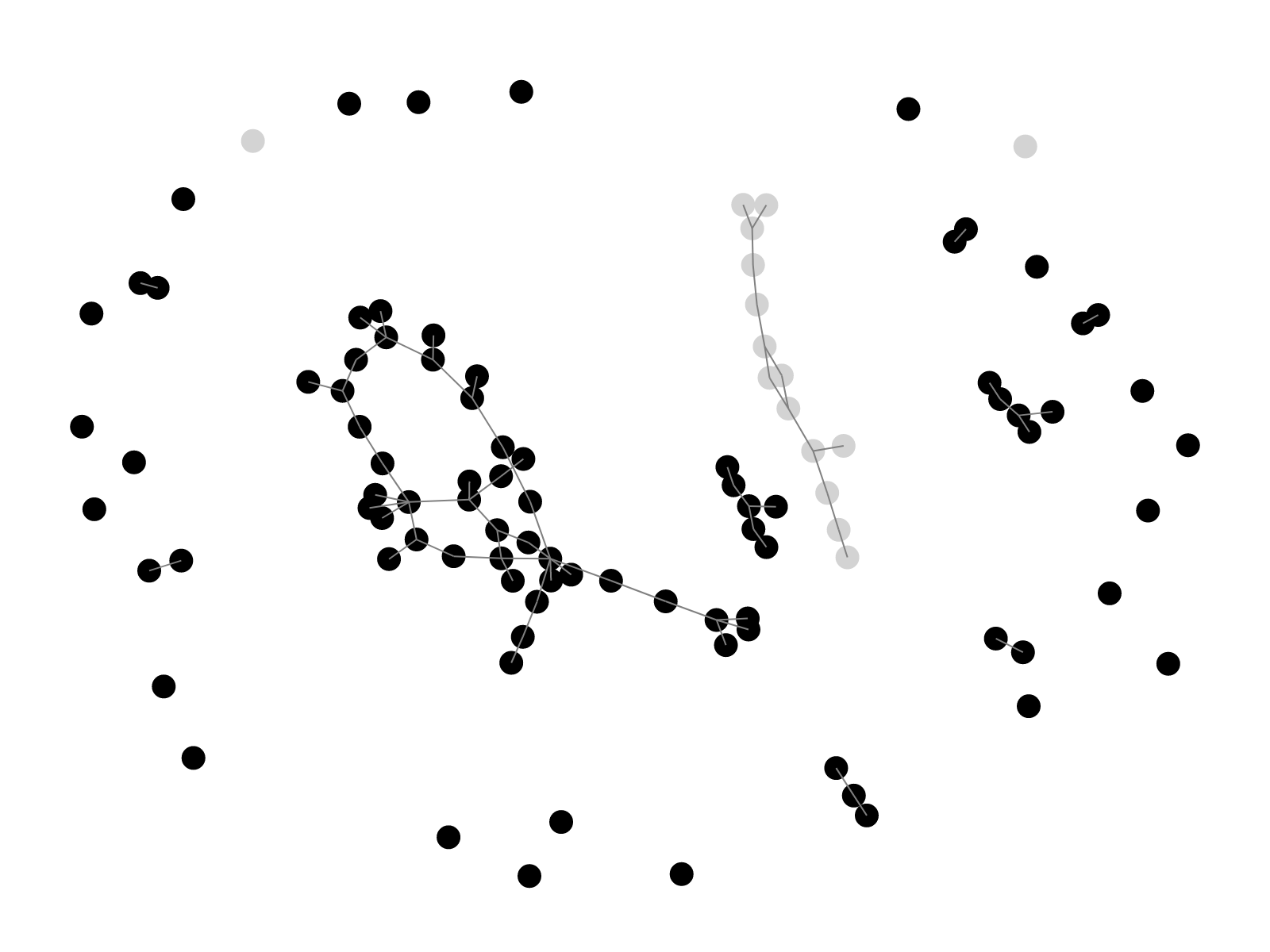}
        \label{fig:Figure2}
    \end{subfigure}
    \caption{Two visual representations of the same graph: (Left) one in 3D space with node coordinates and (Right) the other projected in 2D space, underscoring that the graph's inherent properties are defined in a non-Euclidean space.}
    \label{fig:identical_graphs}
\end{figure}

In our exploration of the prediction of mechanical properties from their graphs, a key presumption is that these graph representations sufficiently capture the geometric information of the rocks. However, it is important to note that the graph representations are not inherently embedded in Euclidean space, suggesting that the captured geometric information is necessarily represented in the graph's connectivity. For example, Figure \ref{fig:identical_graphs} shows two visually distinct yet identical graphs. Despite the change in their visual representation, the underlying topological properties remain constant. Although the visualization does not impact the accuracy of graph-based predictions, it may need to be clarified whether the non-Euclidean representations reflect the microstructures accurately enough to predict the mechanical properties. Therefore, in this section, we employ only graph-level topological features without node features to predict the elastic moduli to verify that these graph-based representations are informative for the prediction of mechanical properties independent of their visual interpretations.

As part of our verification process, we employ a random forest regressor model, using graph topological features as inputs, to predict the elastic moduli. The random forest model operates by constructing numerous decision trees during training and outputting the mean prediction of the individual trees for non-linear regression tasks \citep{gromping2009variable}. For the prediction, we extracted six global topological features from each of the solid and pore graph datasets: average number of edges and nodes, vertex degree, closeness centrality, eigenvector centrality and pagerank (These graph metrics are defined in Appendix \ref{section:Appendix_graph_topo_metrics}). Combining the features, we obtain a set of twelve distinct features that encapsulate a global perspective of the rock microstructure. The predictive performance of this model is used as an indicator to gauge the usefulness of graph topological features in determining the mechanical properties of rocks.

\begin{figure}[pos=H]
\centering
\includegraphics[width=\textwidth]{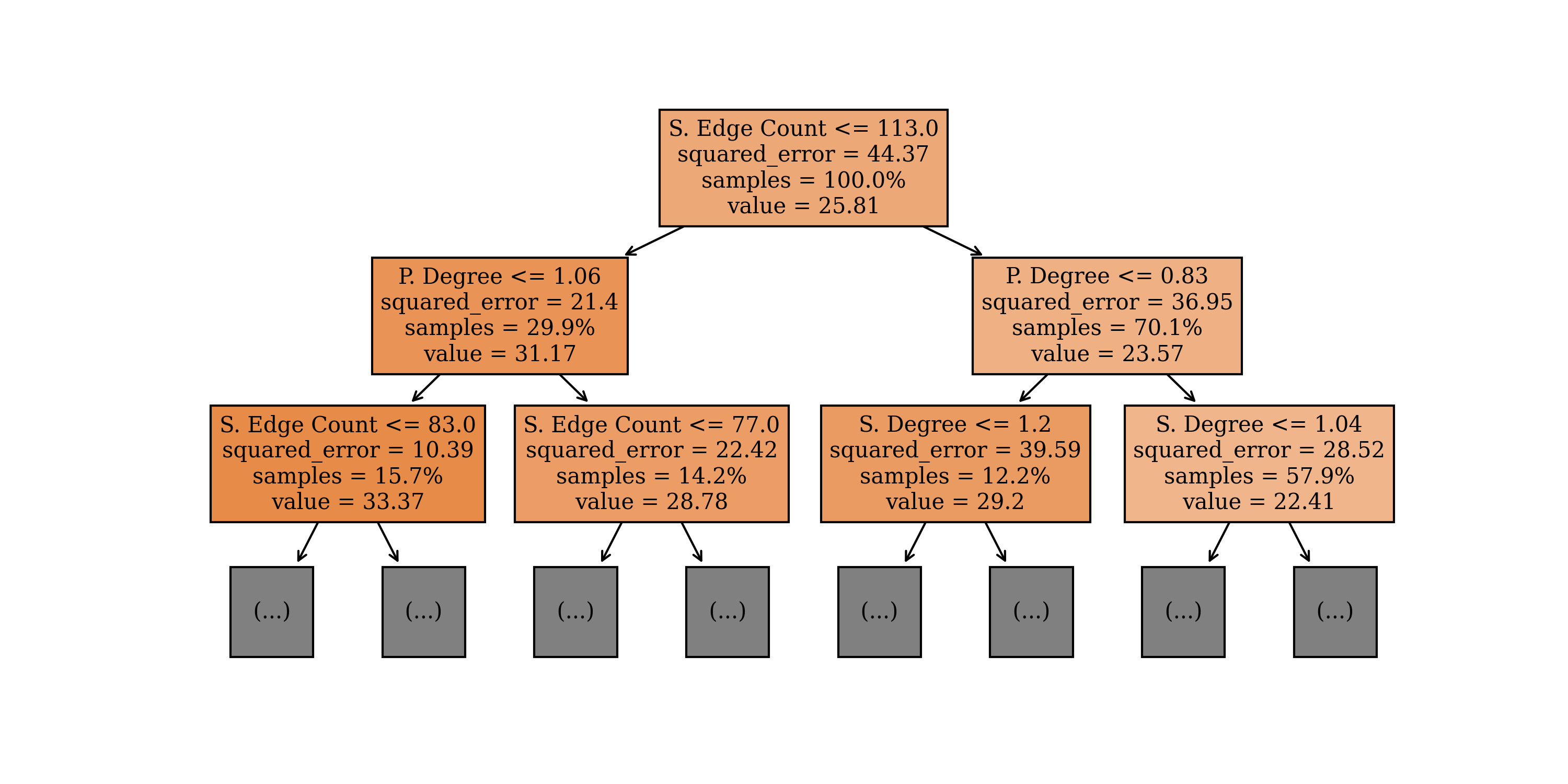}
\caption{An example of a decision tree from the random forest regressor model used to predict Bulk Modulus ($K$). The tree is truncated at a depth of 2 for visualization. Each internal node represents a decision rule, with the condition displayed. In the full model, an ensemble of decision trees contributes to the final prediction by averaging their outputs, enhancing prediction stability and mitigating overfitting through the collective decision-making across varied tree structures.}
\label{fig:TDA_RandomForestRegressor}
\end{figure} 

\begin{figure}
    \centering
    \begin{subfigure}{0.45\textwidth}
        \includegraphics[width=\textwidth]{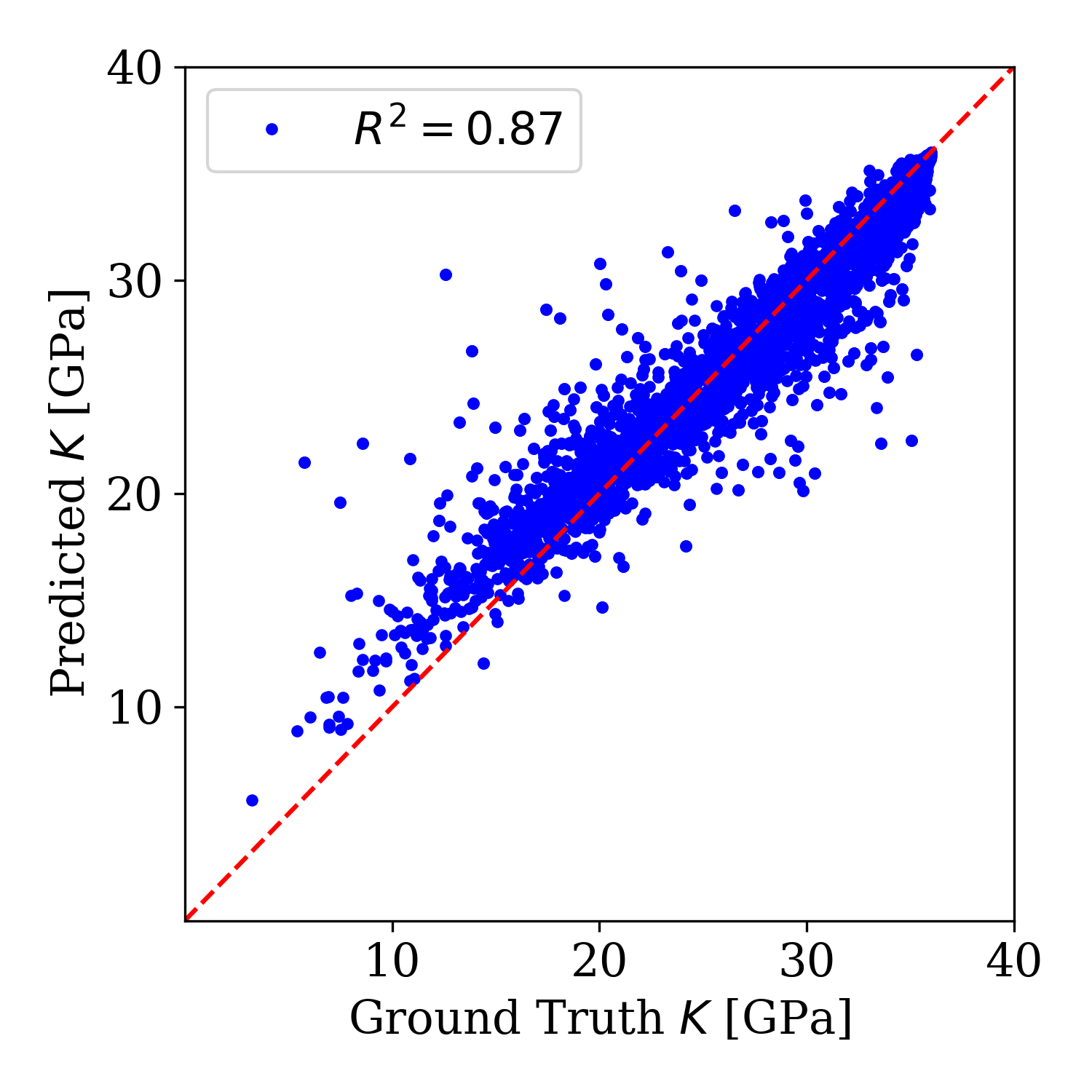}
        \label{fig:random_forest_K}
    \end{subfigure}
    \hfill
    \begin{subfigure}{0.45\textwidth}
        \includegraphics[width=\textwidth]{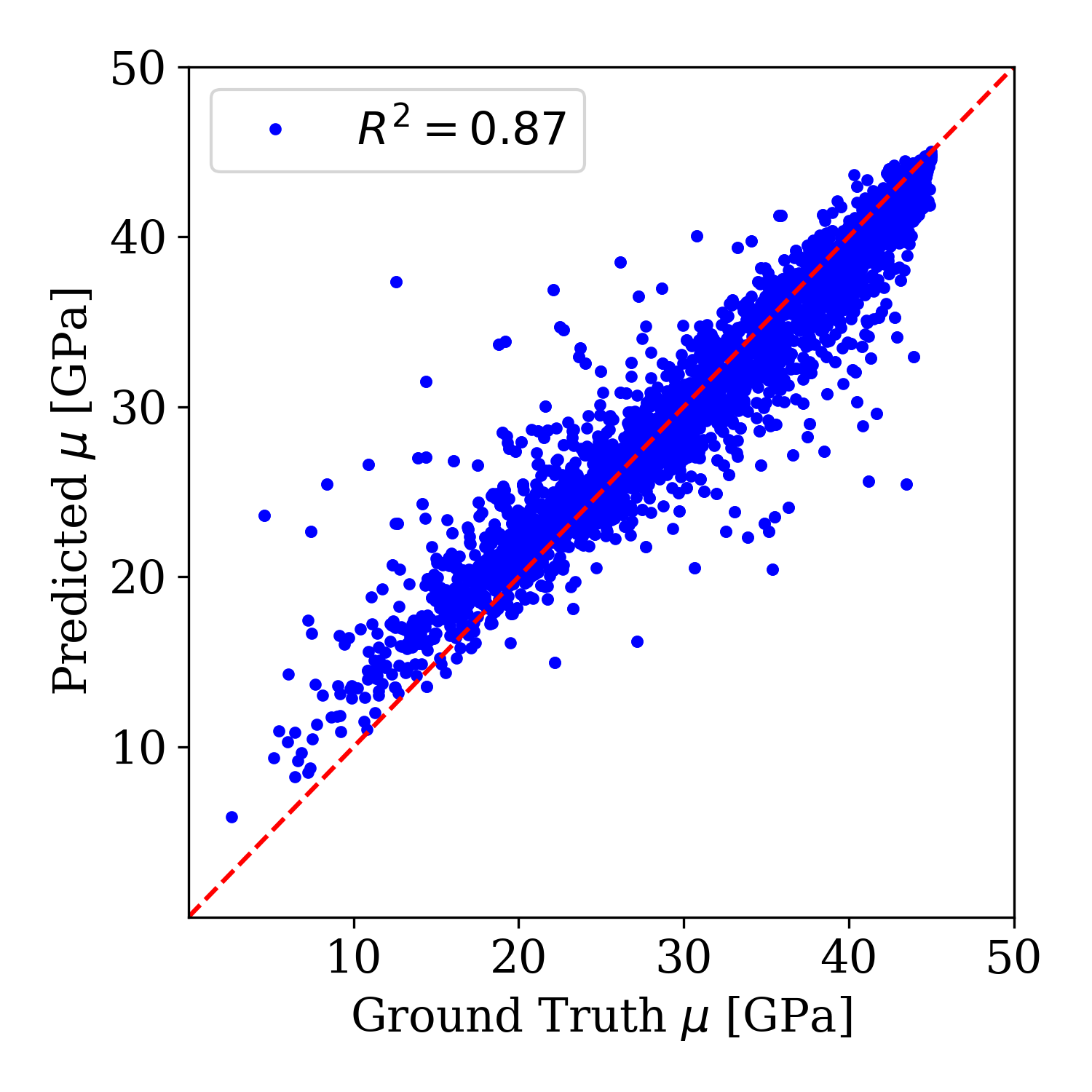}
        \label{fig:random_forest_mu}
    \end{subfigure}
    \caption{Random forest regressor prediction results (Left: bulk moduli prediction, Right: shear moduli prediction)}
    \label{fig:TDA_randomforest_prediction}
\end{figure}

\begin{figure}[pos=H]
\centering
\includegraphics[width=0.5\textwidth]{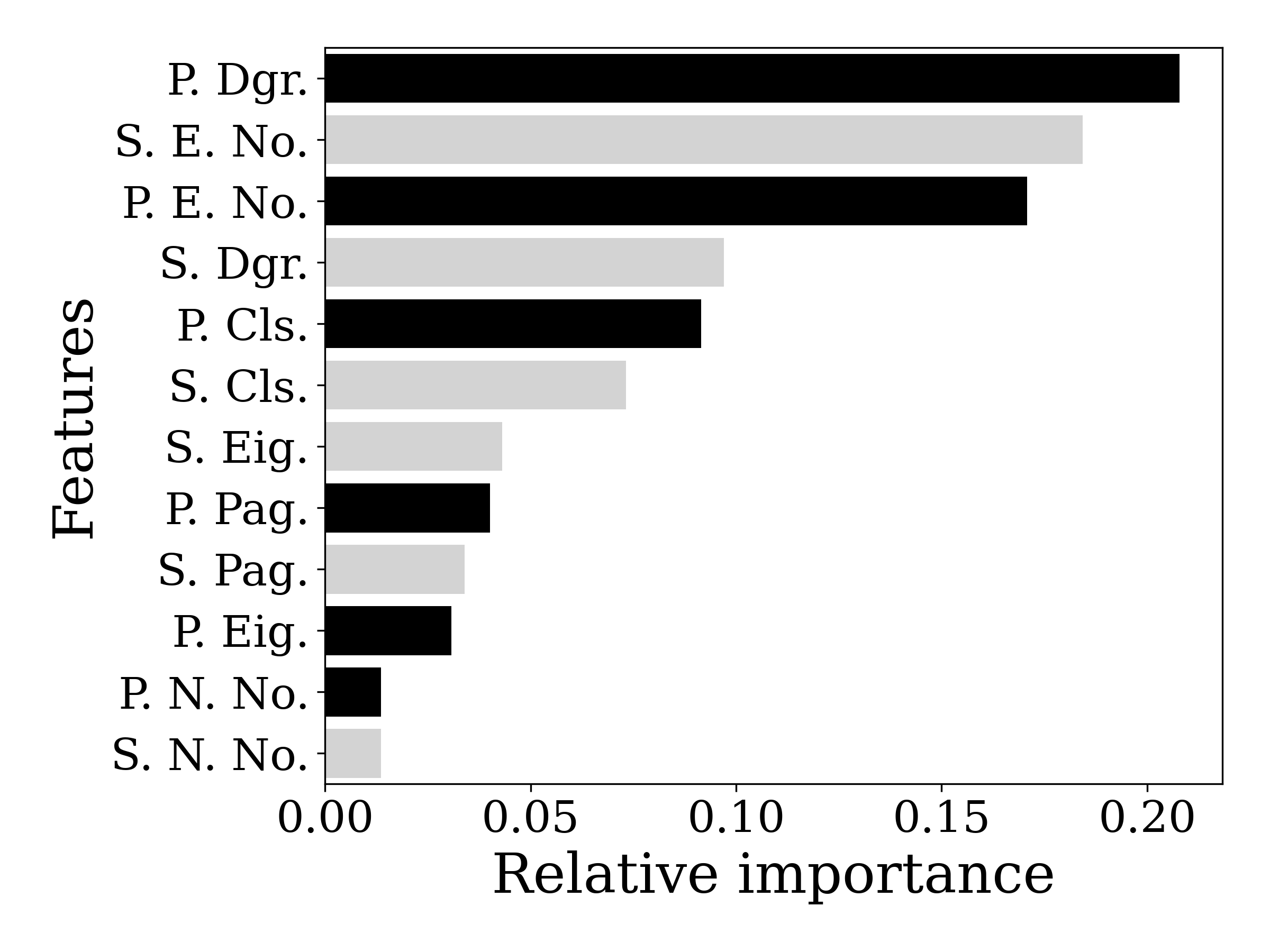}
\caption{Relative importance of different features in the prediction of the elastic moduli using the random forest model. Light gray bars represent solid features, while black bars represent pore features. Feature names are abbreviated for clarity: 'S.' and 'P.' refer to 'Solid' and 'Pore', respectively. 'N. No.': 'Node Number'; 'E. No.': 'Edge Number'; 'Dgr.': 'Degree'; 'Cls.': 'Closeness Centrality'; 'Eig.': 'Eigenvector Centrality'; 'Pag.': 'Pagerank'.}
\label{fig:TDA_Feature_Importance}
\end{figure}

The dataset was divided into training (80 \%), testing (10 \%), and validation (10 \%) subsets. Figure \ref{fig:TDA_RandomForestRegressor} represents one decision tree from the total 50-decision tree in the random forest model to predict the moduli. Each decision tree makes a decision by decreasing the variance of the model and the predicted values are the mean values of the total decision trees given data. Figure \ref{fig:TDA_randomforest_prediction} shows the prediction results for the test dataset from the decision tree model. The random forest regressor model, trained using global topological features, achieved $R^2$ scores of 0.87 for predicting the bulk and the shear modulus. In addition, we quantify the importance of topological features to identify the significant factors to determine the rock property using the Mean Squared Error (MSE) magnitude for the regression. Figure \ref{fig:TDA_Feature_Importance} shows the importance of topological features in the prediction of the elastic moduli. The node degree in pore graph emerged as the most significant determinant. This metric directly relates to pore connectivity within the rock's microstructure. Such features play a fundamental role in stress transmission across the rock matrix. Following this, the number of edges in both the solid and pore graphs occupy the subsequent positions in importance. These attributes are indicative of a dense interlinking within the respective phases. This interconnectedness, as evidenced by the significance of the topological features, influences how stress is distributed within the rock matrix. While in some scenarios it might lead to a more uniform distribution contributing to increased rigidity and resistance against deformation, in others, it could imply potential weak points or pathways for stress concentration, leading to areas of reduced stiffness.

\begin{figure}
    \centering
    \begin{subfigure}{0.45\textwidth}
        \includegraphics[width=\textwidth]{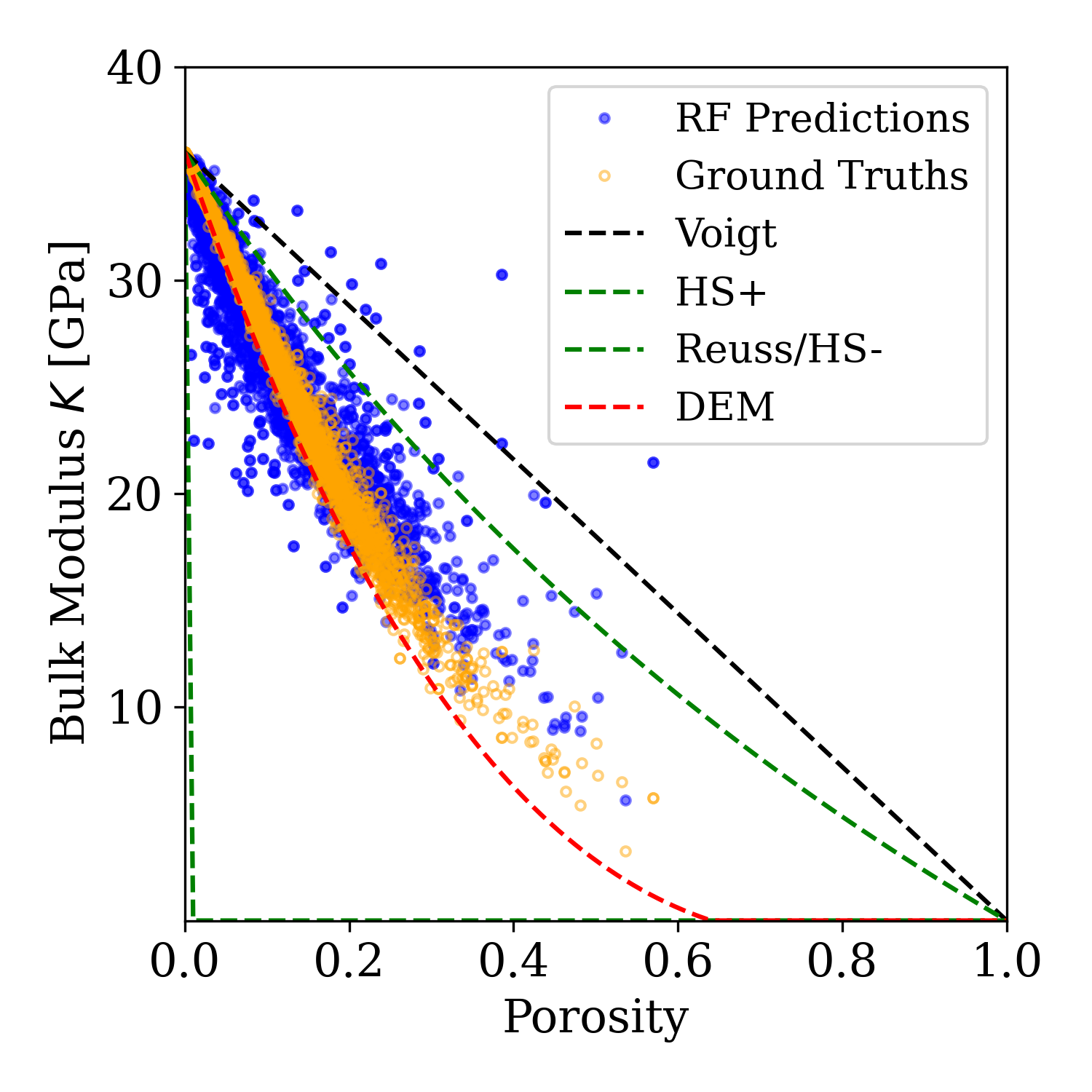}
        \label{fig:random_forest_K_bounds}
    \end{subfigure}
    \hfill
    \begin{subfigure}{0.45\textwidth}
        \includegraphics[width=\textwidth]{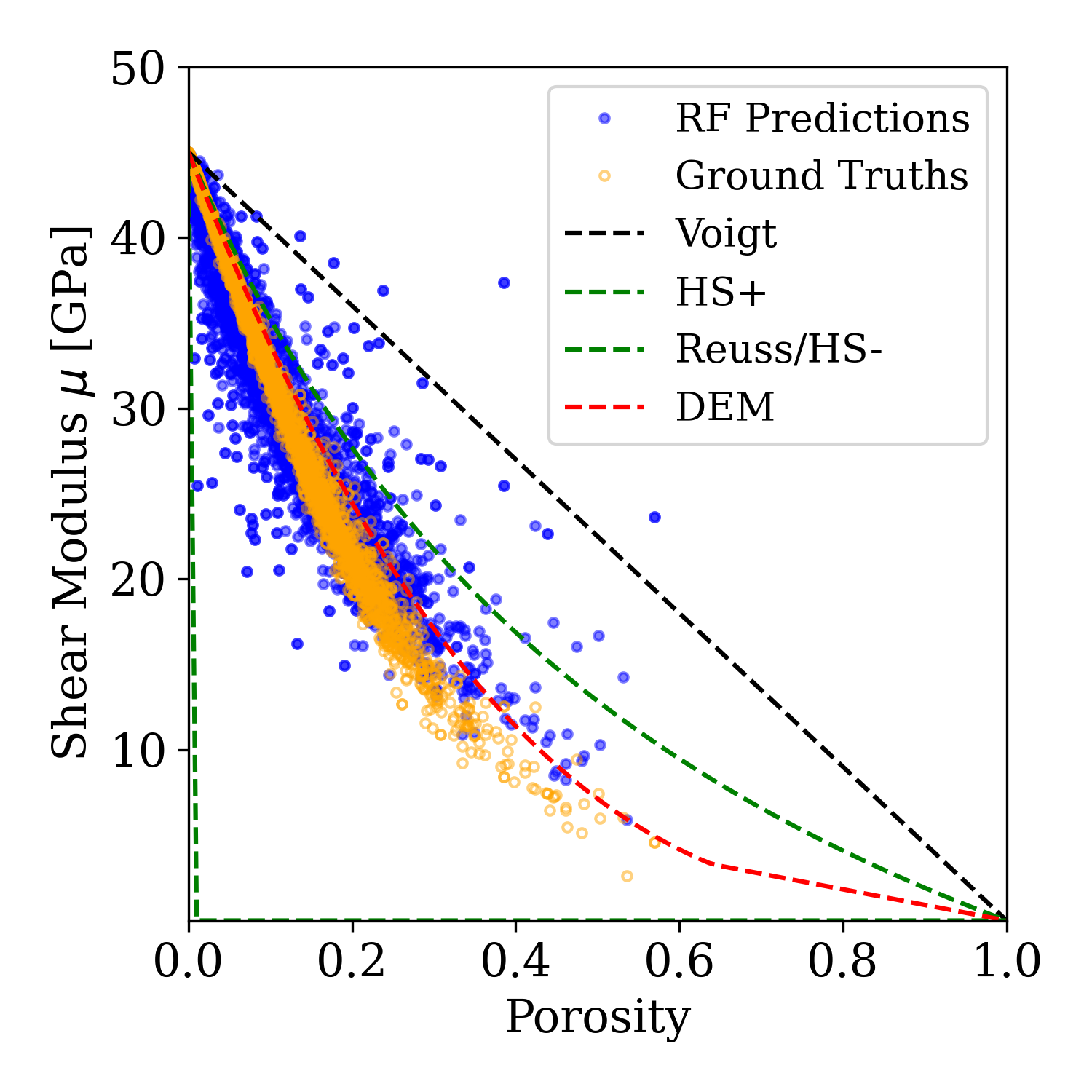}
        \label{fig:random_forest_mu_bounds}
    \end{subfigure}
    \caption{Predictions of the random forest regressor for bulk (left) and shear (right) moduli, compared with theoretical bounds and computed moduli across varying porosities. 'HS' refers to Hashin-Shtrikman bounds, and 'DEM' to Differential Effective Medium. 'Ground Truths' refer to moduli computed directly on the digital rock subcube using full-physics numerical simulations.}
    \label{fig:TDA_randomforest_prediction_bounds}
\end{figure}

We also employ the differential effective medium (DEM) models \citep{norris1985effective, berryman1980long}, and traditional bounds including Voigt-Reuss \citep{reuss1929berechnung} and Hashin-Shtrikman \citep{hashin1963variational} for assessing the qualitative performance of the graph-based predictions. For a fair comparison, we tuned the geometrical parameters of the DEM model with the identical training dataset. The DEM model achieved an $R^2$ of $0.93$ for both bulk and shear moduli as shown in Figure \ref{fig:DEM_test_results} (Please see the Appendix \ref{section:Appendix_DEM_results} for the details). While the DEM results are better than the random forest model ($R^2$ of 0.87), the potentials of the graph-based predictions are notable given that the random forest model is based solely on the graph's connectivity, without incorporating porosities or node features in this analysis. Figure \ref{fig:TDA_randomforest_prediction_bounds} shows that the random forest predictions are scattered around the ground truths yet closely align with the optimized DEM model within the Voigt-Reuss and Hashin-Shtrikman bounds. Specifically, 99 \% and 97 \% of the predicted values fall within the upper and lower limits of the Voigt-Reuss and Hashin-Shtrikman bounds, respectively.

Our observations provide support to our initial premise that the topological features of a graph, representing the geometric structure of rock microstructures, contain meaningful information for the prediction of rock's elastic moduli. The results also hint at the potential of these topological features for enhancing the predictive performance of machine learning models aimed at property prediction tasks. In subsequent sections, we extend this investigation to incorporate GNNs, leveraging both the topological features and node features of the graphs to improve prediction performance further.

\subsection{Prediction performance of the GNN}
The graphs created by the mapper algorithm were used to improve the predictability of elastic moduli of digital rocks by employing a Graph Neural Network (GNN) model. The training dataset consisted of four rock types (B1, B2, FB1, and FB2) and multiple subcube sizes (90, 100, 180). This variability emphasizes the advantage of the GNN over traditional Convolutional Neural Networks (CNNs), which are constrained by fixed-size inputs, while GNNs can process graphs of varying sizes \citep{scarselli2008graph}. The dataset was partitioned into training, validation, and testing subsets in a 80:10:10 ratio. A detailed illustration of the GNN model architecture is available in Section \ref{section:GNN_architecture}.

\subsubsection{Tuning Mapper algorithm and GNN Hyperparameters for Optimal Performance}
Initial exploration involved identifying optimal parameters for the Mapper algorithm, specifically the cover interval and the overlap, which influence the resolution and the connectivity of the graphs, respectively. A systematic examination was conducted with cover intervals $\in \{10, 20, 30\}$ and overlaps $\in \{0.10, 0.30, 0.50\}$. The results showed high predictability of the GNN model, with $R^2$ values exceeding $0.972$, as summarized in Table \ref{tab:Mapper_optimization}. Notably, the combination of $10$ for cover intervals and $0.50$ for overlap demonstrated the highest performance ($R^2=0.99$), and were thereby identified as the optimal parameters for our graph-based representation of the dataset.

\begin{table}
\centering
\caption{$R^2$ values from test set for bulk and shear moduli depending on the cover interval and overlap in the mapper algorithm. Each cell contains two $R^2$ values: the first is for the bulk modulus and the second is for the shear modulus.}
\label{tab:Mapper_optimization}
\begin{tabular}{ |c||c|c|c|c|} 
 \hline
 Cover interval / Overlap & $O_1 (0.10)$ & $O_2 (0.30)$ & $O_n (0.50)$ \\ 
 \hline 
 \hline
$C_1 (10)$ & $0.982,  0.980$ & $0.983, 0.982$ & $0.991, 0.992$ \\
\hline
$C_2 (20)$ & $0.973, 0.974$ & $0.988, 0.987$ & $0.990, 0.988$ \\
\hline
$C_3 (30)$ & $0.979, 0.981$ & $0.982, 0.983$ & $0.985, 0.984$ \\
\hline
\end{tabular} 
\end{table}

\begin{figure}
\centering
\includegraphics[width=0.5\textwidth]{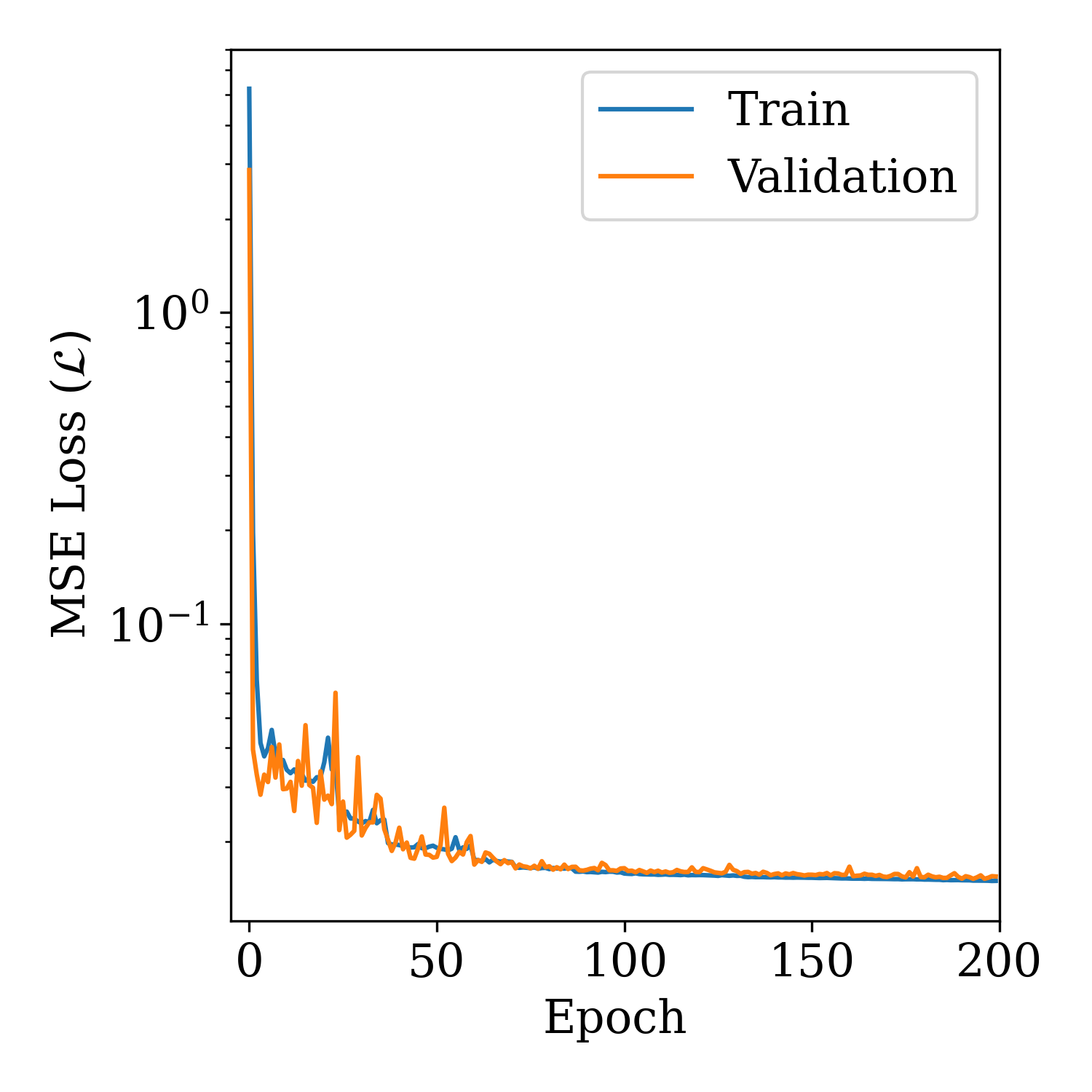}
\caption{Variation of mean square error (MSE) loss with epoch during GNN training}
\label{fig:GNN_R2_epoch}
\end{figure}

\begin{figure}
    \centering
    \begin{subfigure}{0.45\textwidth}
    \centering
        \includegraphics[width=\textwidth]{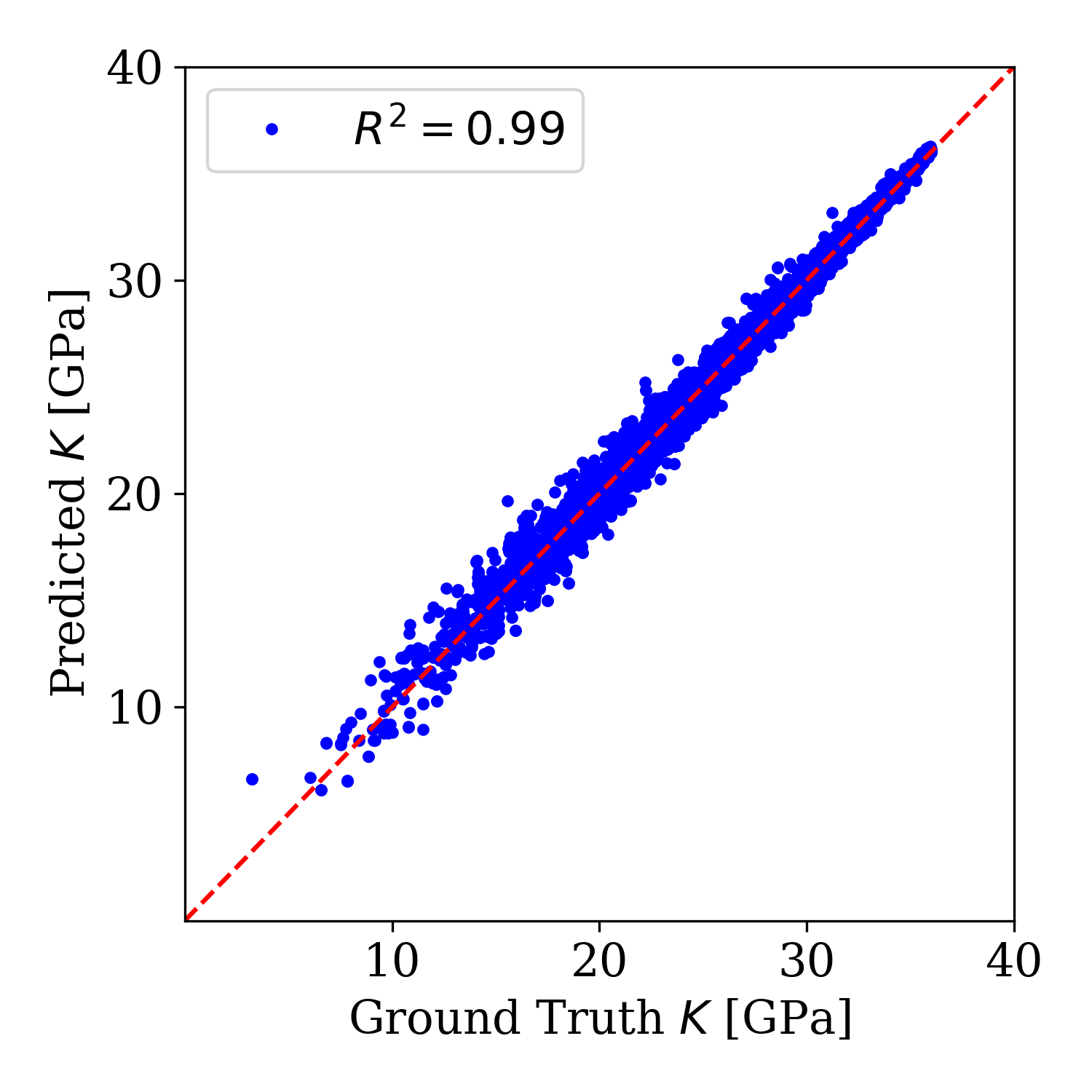}
        \caption{Bulk moduli prediction}
    \end{subfigure}
    \hspace{0.1cm}
    \begin{subfigure}{0.45\textwidth}
    \centering
        \includegraphics[width=\textwidth]{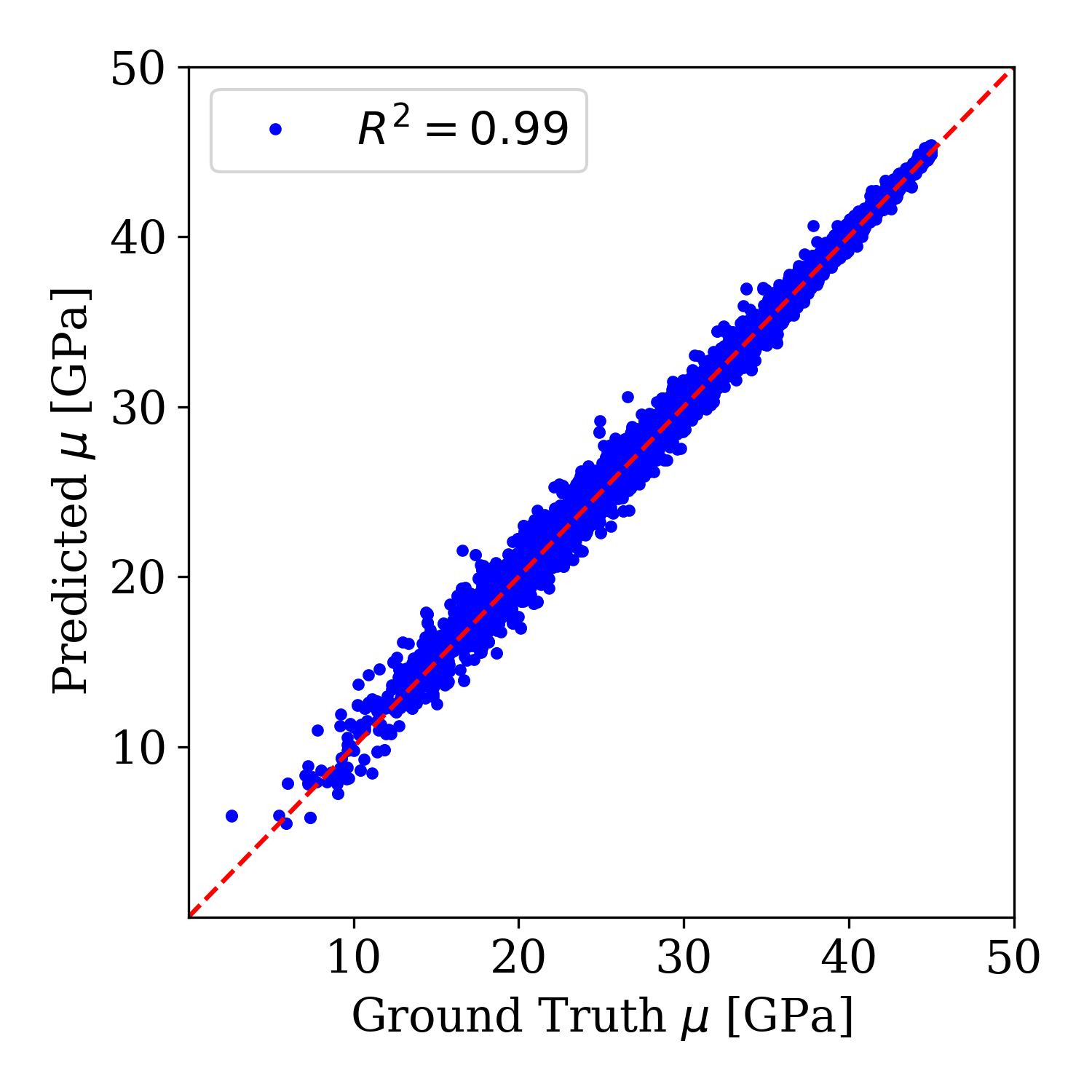}
        \caption{Shear moduli prediction}
    \end{subfigure}
    \caption{GNN testing results for network trained with 4 rocks (B1, B2, FB1, FB2) and three subcube sizes (90, 100, 180)}
    \label{fig:GNN_training_results}
\end{figure}

\begin{figure}
    \centering
    \begin{subfigure}{0.45\textwidth}
        \includegraphics[width=\textwidth]{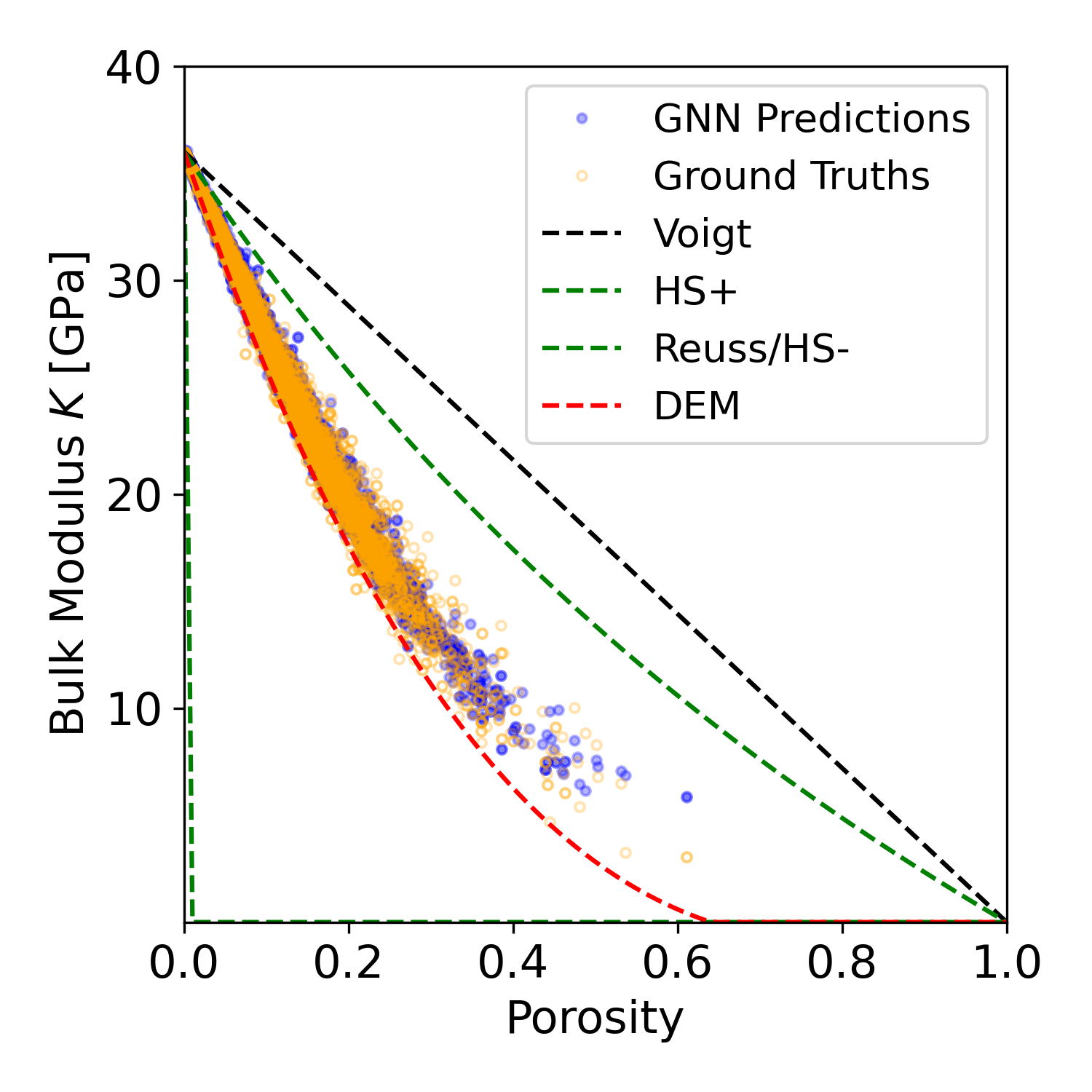}
        \label{fig:random_forest_K_bounds}
    \end{subfigure}
    \hfill
    \begin{subfigure}{0.45\textwidth}
        \includegraphics[width=\textwidth]{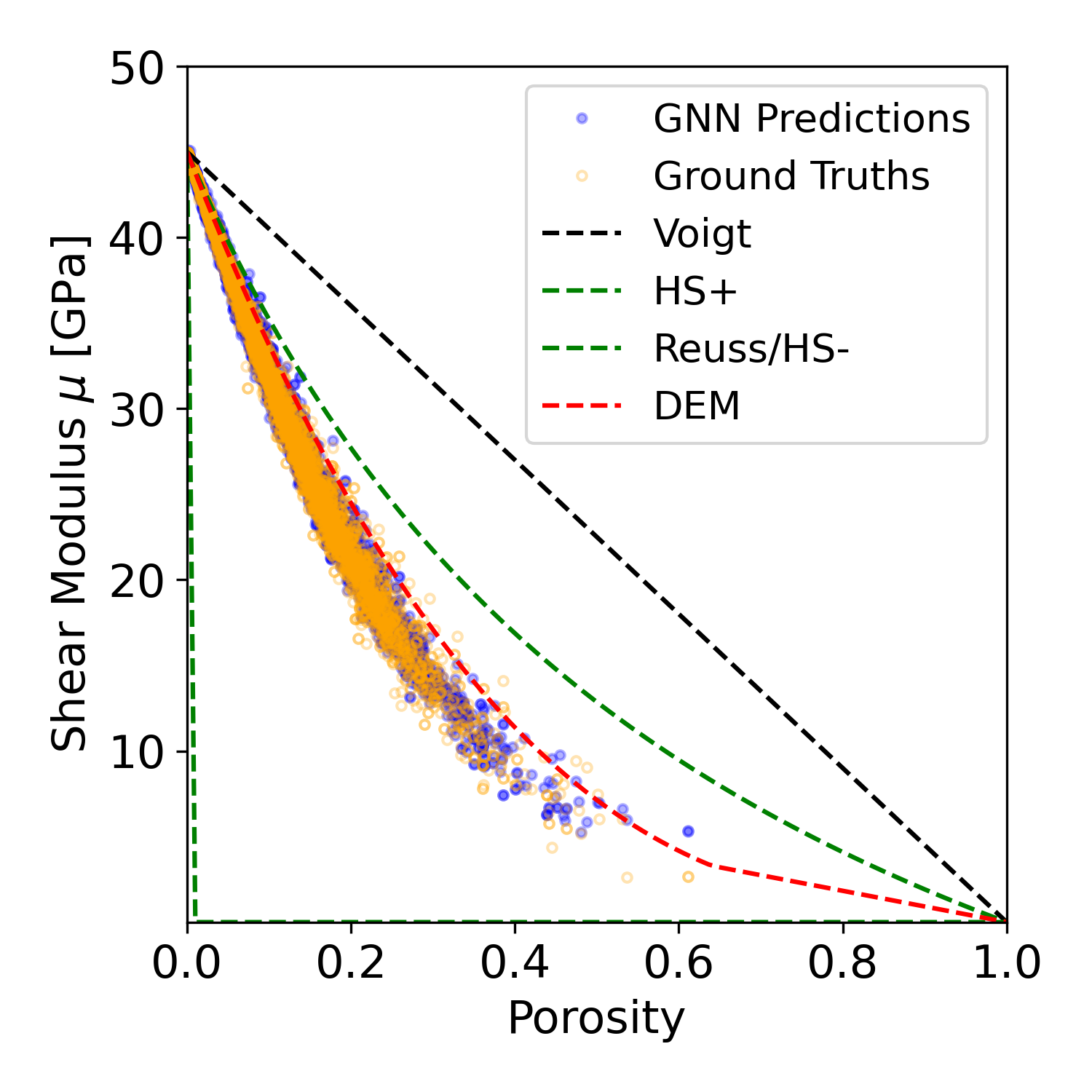}
        \label{fig:random_forest_mu_bounds}
    \end{subfigure}
    \caption{GNN prediction results for bulk (left) and shear (right) moduli, compared with theoretical bounds and computed moduli across varying porosities. 'HS' refers to Hashin-Shtrikman bounds, and 'DEM' to Differential Effective Medium. 'Ground Truths' refer to moduli computed directly on the digital rock subcube using full-physics numerical simulations.}
    \label{fig:GNN_prediction_bounds}
\end{figure}

Next, we proceeded with hyperparameter tuning of the GNN model, focusing on the batch size $\in \{2^5, 2^7, 2^9, 2^{11}\}$ and the dropout ratio $\in \{0, 0.5\}$. Note the wide range flexibility of batch size in the GNNs, which we will further discuss in the following Section \ref{section:GNN_vs_CNN}. Table \ref{tab:Hyperparameter_tuning} provides the performance of the model for each set of parameters. According to our results, a batch size of $2^9$ and a dropout of $0.5$ optimized the performance of the GNN. The model's mean square error (MSE) loss decreased over epochs, plateauing around the $200 th$ epoch, as shown in Figure \ref{fig:GNN_R2_epoch}. The comparable values of training and validation MSE imply that the model did not overfit. The model's predictive ability is indicated by the $R^2$ value of 0.99 on the testing set (Figure \ref{fig:GNN_training_results}). In addition, Figure \ref{fig:GNN_prediction_bounds} shows the GNN predictions almost overlap with ground truths, and all the results fall between theoretical bounds.
This result suggests that the GNN model can predict elastic moduli accurately based on the mapper-constructed graph from two sandstones (Berea and Fontainebleu sandstone) and different subcube image sizes.

\begin{table}
\centering
\caption{Performance of the GNN model for each set of parameters (batch size and dropout ratio). The reported $R^2$ values are from the test set.}
\label{tab:Hyperparameter_tuning}
\begin{tabular}{|c|c|c|c|} 
 \hline
 Batch Size & Dropout Ratio & $R^2$ ($K$) & $R^2$ ($\mu$) \\ 
 \hline
 $2^5$ & 0 & 0.983 & 0.984 \\
 \hline
 $2^5$ & 0.5 & 0.985 & 0.982 \\
  \hline
 $2^7$ & 0 & 0.984 & 0.984 \\
 \hline
 $2^7$ & 0.5 & 0.988 &0.990 \\
\hline
 $2^9$ & 0 & 0.988 & 0.990 \\
 \hline
 $2^9$ & 0.5 & 0.991 &0.992 \\
\hline
 $2^{11}$ & 0 & 0.990 & 0.989 \\
 \hline
 $2^{11}$ & 0.5 & 0.990 &0.984 \\
 \hline
\end{tabular} 
\end{table}

\subsubsection{Assessing Predictive Generalizability on Unseen Rocks}
We evaluate the ability of our model to predict the properties of unseen rock samples, thus demonstrating its generalizability. The capacity of a deep learning model to accurately predict outcomes on the unseen data is an essential measure of model performance \citep{kawaguchi2016deep}. This process involves testing the model on data that significantly differs from the training data, thereby verifying whether the model can generalize patterns learned during the training phase. For digital rocks, this signifies testing the model on unseen rock types with differing microstructural characteristics. We evaluated our model's generalization capability by testing it on an unseen rock type, the Castlegate (CG) sandstone. This rock type exhibits distinct microstructural properties in comparison to the Berea and Fontainebleau sandstones used in the training process. In our dataset, Berea sandstone show porosities of 16.5 \% for the B1 sample and 19.6 \% for the B2 sample, and the porosities of the Fontainebleau samples are 9.2 \% for FB1 and 3.4 \% for FB2. By comparison, the Castlegate sample (CG) has a porosity of 22.2 \%. Previous studies also reported the mean grain size of Berea sandstone is $51 \mu m$ \citep{safari2021characterization}, and the Fontainebleau sandstone's is $250 \mu m$ \citep{gomez2010laboratory}. In contrast, the Castlegate sandstone has mean grain size of $64 \mu m $ \citep{safari2021characterization}.

\begin{figure}[pos=h!]
\centering
\begin{subfigure}[b]{0.45\textwidth}
\centering
\includegraphics[width=\textwidth]{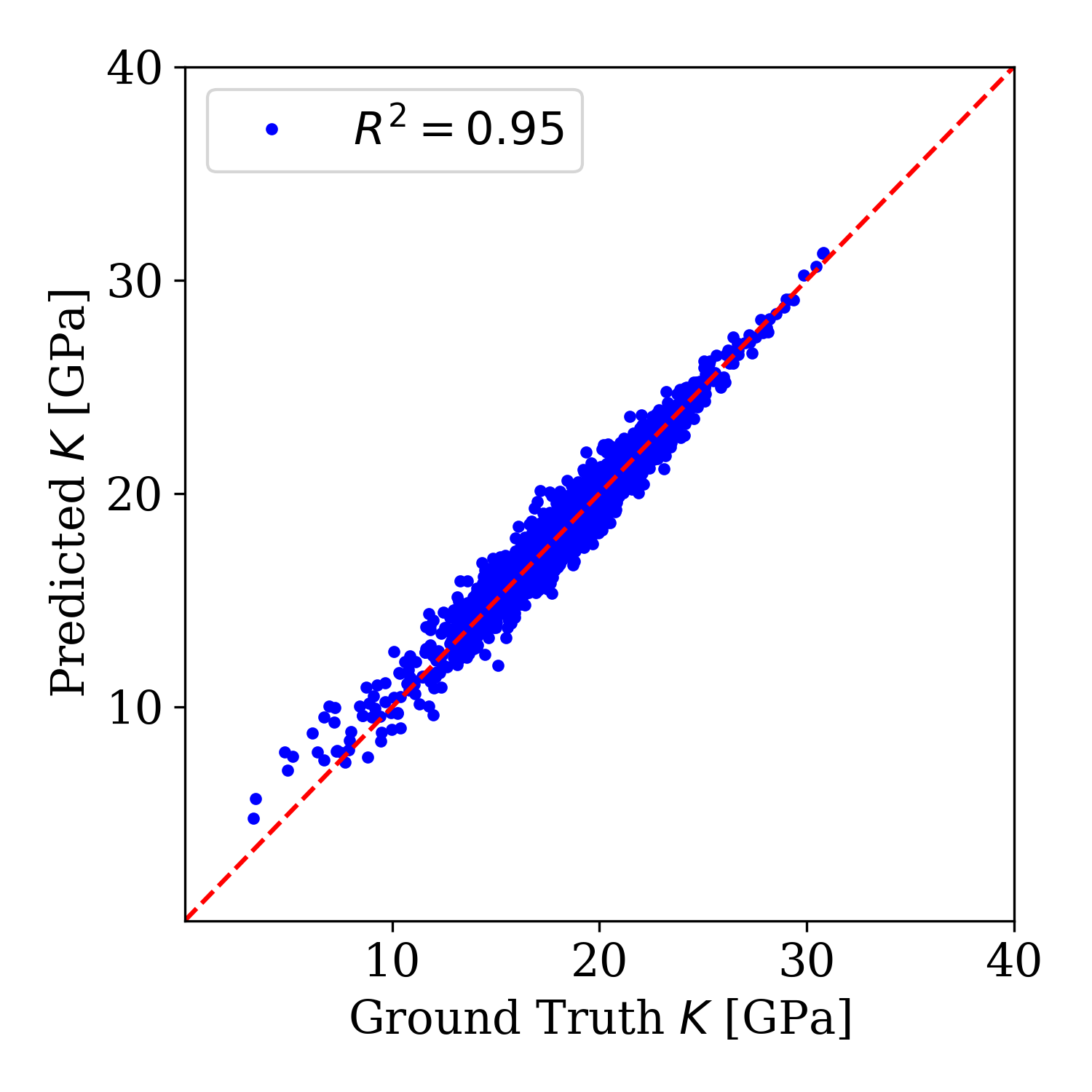}
\caption{Bulk modulus predictions}
\label{fig:bulk_modulus_prediction}
\end{subfigure}
\hspace{0.1cm}
\begin{subfigure}[b]{0.45\textwidth}
\centering
\includegraphics[width=\textwidth]{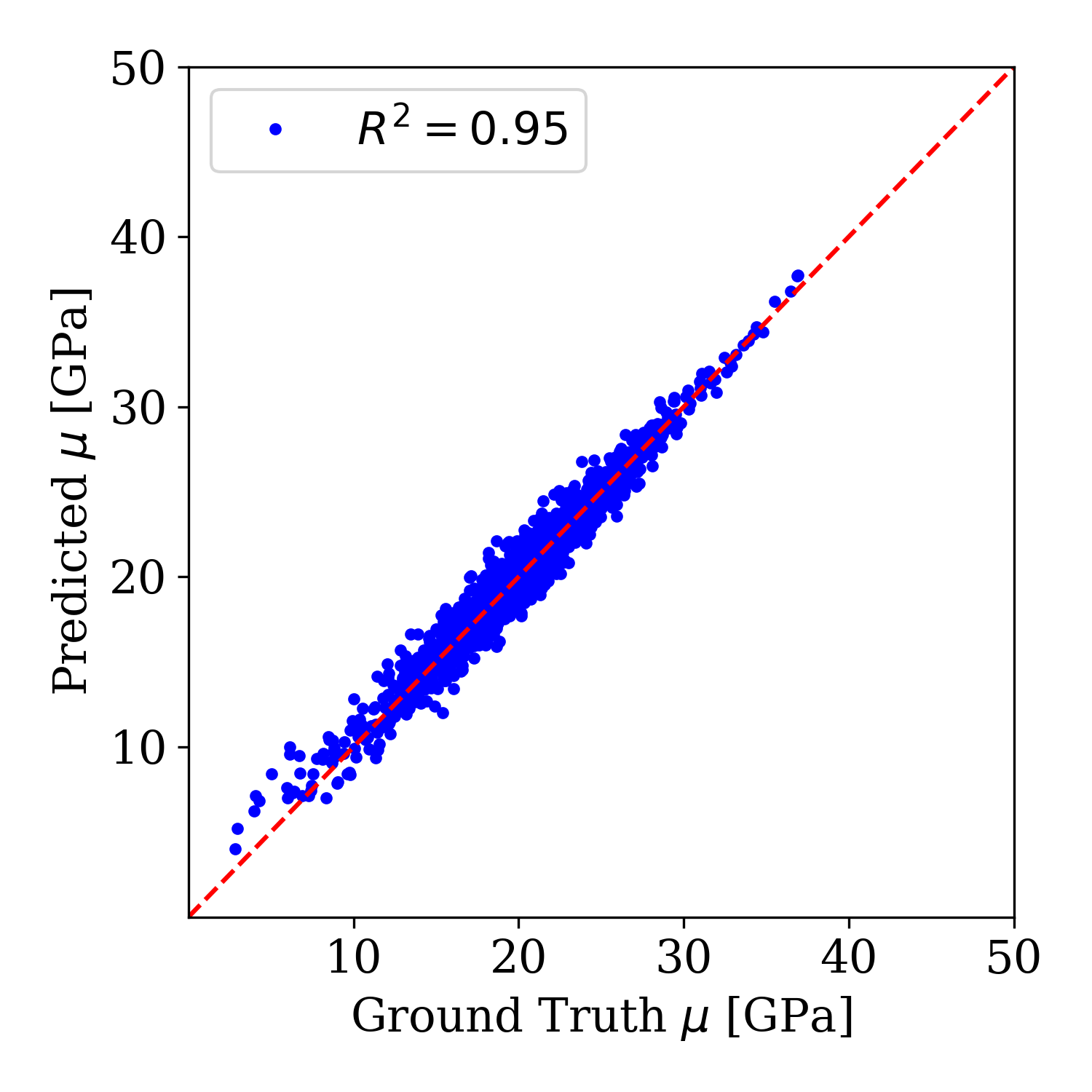}
\caption{Shear modulus predictions}
\label{fig:shear_modulus_prediction}
\end{subfigure}
\caption{Predictive results of the GNN model for the unseen Castlegate sandstone. The model was trained with four rock types (B1, B2, FB1, FB2) and three subcube sizes (90, 100, 180).}
\label{fig:GNN_Unseen_rock_prediction_results}
\end{figure}

In spite of being trained exclusively on Berea and Fontainebleau sandstones, the GNN model achieved an $R^2$ value of 0.95 for both bulk and shear moduli predictions when tested on the unseen Castlegate sandstone, as shown in Figure \ref{fig:GNN_Unseen_rock_prediction_results}. This high accuracy in an unseen context underlines the model's strong generalizability. It suggests that the learned features and patterns are not overly specific to the training rock types, but have captured general underlying structures applicable to a broader variety of sandstones \citep{nguyen2015deep}.

\subsubsection{Property Prediction for Unseen Subcube Sizes}
In addition to unseen rock types, we assessed the model's capability to predict elastic moduli of graphs constructed from unseen subcube image sizes (subcube image size 150). The field of digital rock analysis often encounters varying image sizes in the samples. Factors such as the precision of the imaging technique or the specific area of interest within the rock \citep{blunt2013pore} can influence the size of the analyzed subcube. This introduces variability in the voxel size of the input data. Most traditional deep learning models, including CNNs, require fixed-size inputs. A change in voxel size necessitates retraining of the model, which can be computationally expensive and time-consuming, negating some of the computational efficiency advantages of deep learning. However, GNNs present a significant advantage in this aspect. Owing to their inherent design, GNNs can handle inputs of varying sizes, thereby removing the need for model retraining when the input size changes \citep{bronstein2017geometric}.

\begin{figure}[pos=h]
\centering
\begin{subfigure}[b]{0.45\textwidth}
\centering
\includegraphics[width=\textwidth]{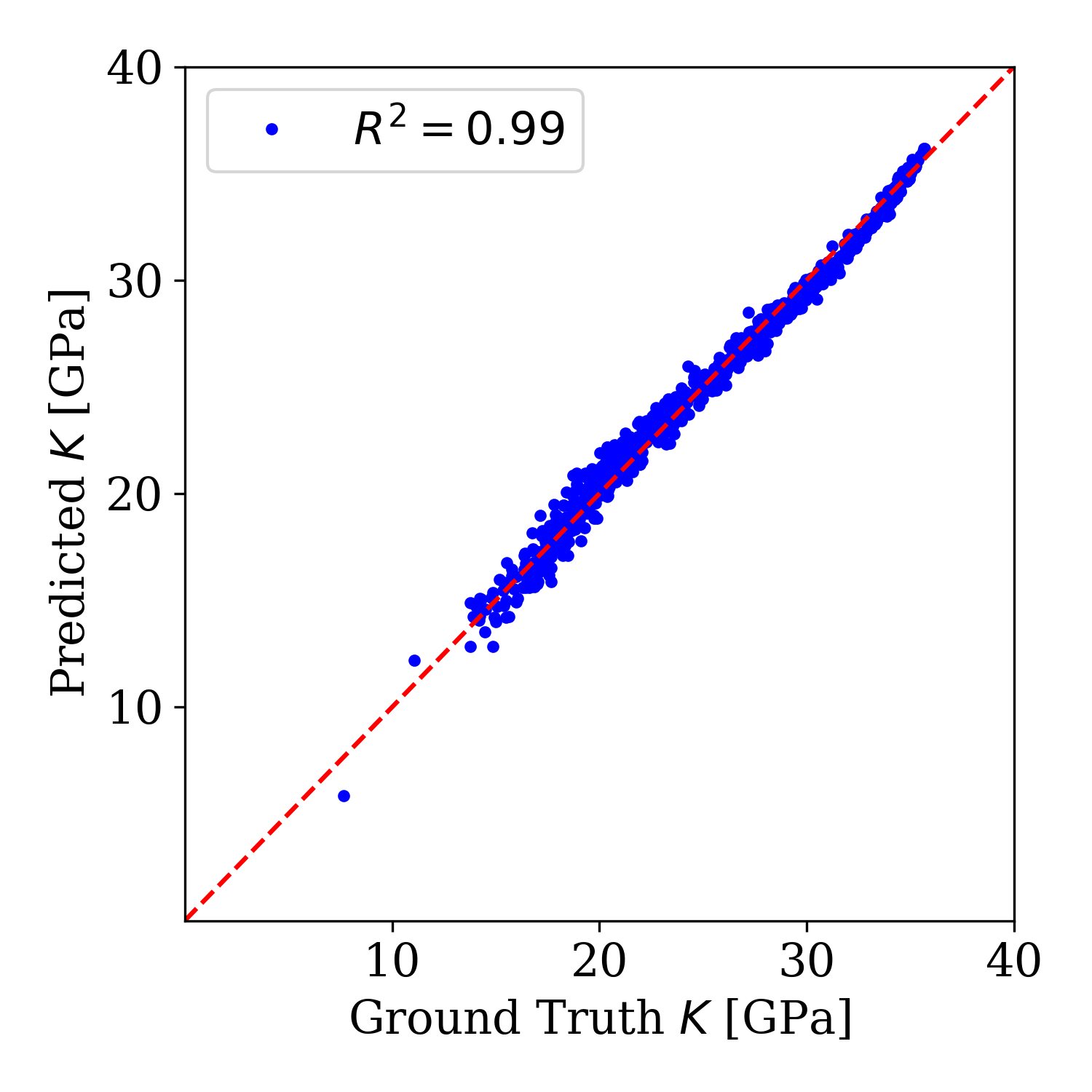}
\caption{Bulk modulus predictions}
\label{fig:bulk_modulus_prediction_unseen}
\end{subfigure}
\hspace{0.1cm}
\begin{subfigure}[b]{0.45\textwidth}
\centering
\includegraphics[width=\textwidth]{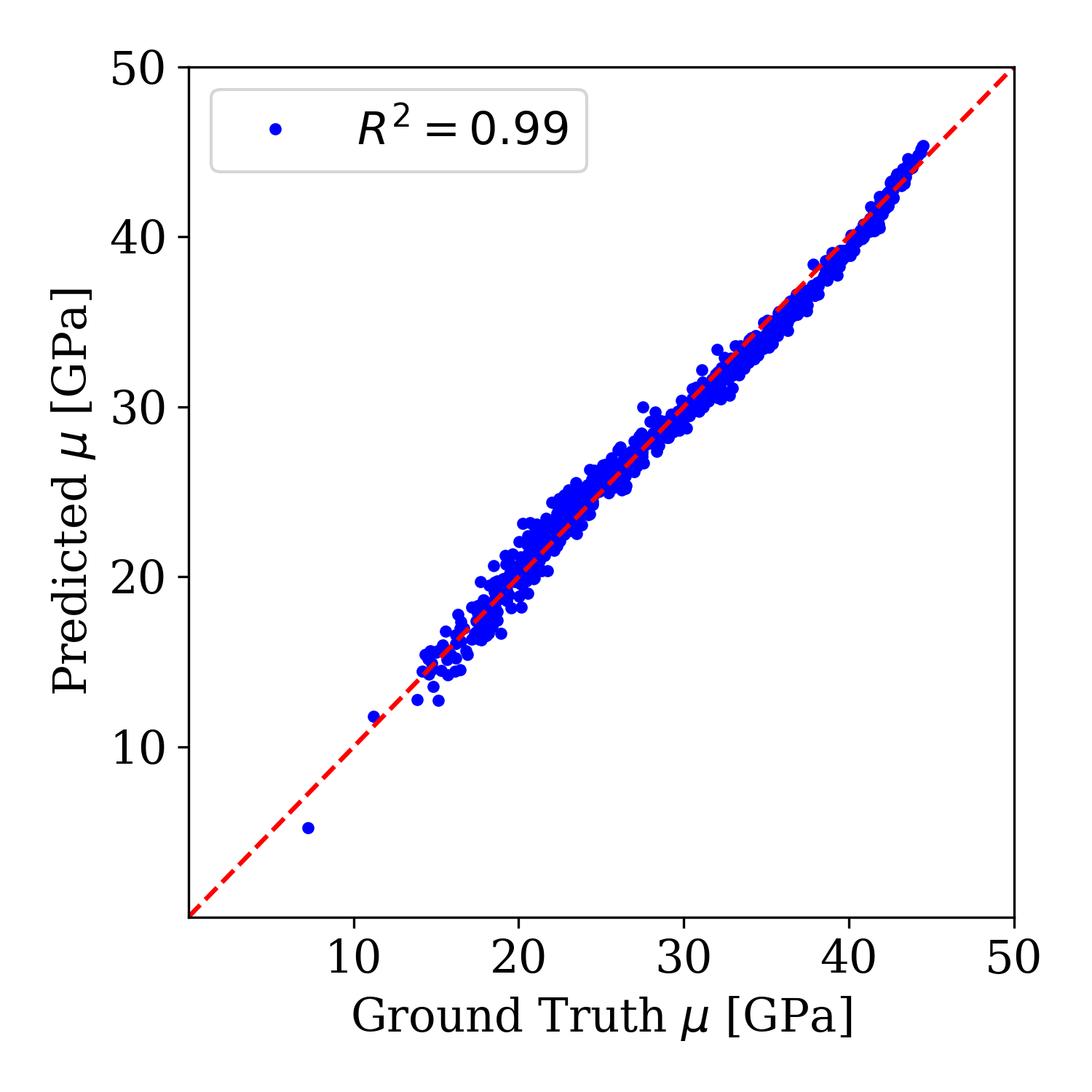}
\caption{Shear modulus predictions}
\label{fig:shear_modulus_prediction_unseen}
\end{subfigure}
\caption{GNN model prediction results for unseen graphs derived from a subcube size of 150, trained on four rock types (B1, B2, FB1, FB2) and three subcube sizes (90, 100, 180).}
\label{fig:GNN_prediction_results_unseen_subcube}
\end{figure}

Our GNN model effectively demonstrated this capability. Although trained on three different subcube sizes, the model yielded an $R^2$ value of 0.99 for both bulk and shear moduli for unseen graphs constructed from a subcube of size 150 (Figure \ref{fig:GNN_prediction_results_unseen_subcube}). This results underscores the superiority of GNNs in managing variable input sizes, suggesting their appropriateness for applications in the digital rock analysis domain where input sizes can substantially vary.

\subsection{Comparative analysis of GNNs and CNNs}
\label{section:GNN_vs_CNN}
In the realm of deep learning, two significant architectures have seen widespread use: Convolutional Neural Networks (CNNs) and Graph Neural Networks (GNNs). An exact comparison between these two architectures can be challenging due to their inherent differences and their design for different types of data. CNNs are primarily suitable for grid-like data, such as images, by applying convolutional filters that effectively extract local features \citep{albawi2017understanding}. In contrast, GNNs are designed to process data represented in graph structures, making them suitable for scenarios where relationships between entities are complex and non-gridlike \citep{wu2020comprehensive}.

For a meaningful comparison, it is crucial to consider various factors such as the different combinations of layer functions, including pooling and dropout layers, which significantly influence the models' performances. To ensure a fair comparison, we closely mirrored the CNN to the GNN concerning the number of convolutional layers, filter depths of each convolution, and the structure of the multi-layer perceptron (MLP) decoders. We adjusted only the first layer dimension in the MLP due to the summation pooling inherent to the Graph Isomorphism Network (GIN) architecture. We provide detailed specifications of the architectures in Figures \ref{fig:GNN_arch} and \ref{fig:CNN_arch} and Tables \ref{tab:Model_Comparison}. Additionally, to ensure fairness, we tuned the CNN's batch size and dropout ratio as per the GNN. The optimal model results are included in the Appendix \ref{section:Appendix_cnn_results}. In short, the CNN shows its capability to predict effective moduli of subcube size 90 with the $R^2$ of 0.98. 

\begin{figure}
    \centering
    \begin{subfigure}{0.45\textwidth}
        \centering\includegraphics[width=\textwidth]{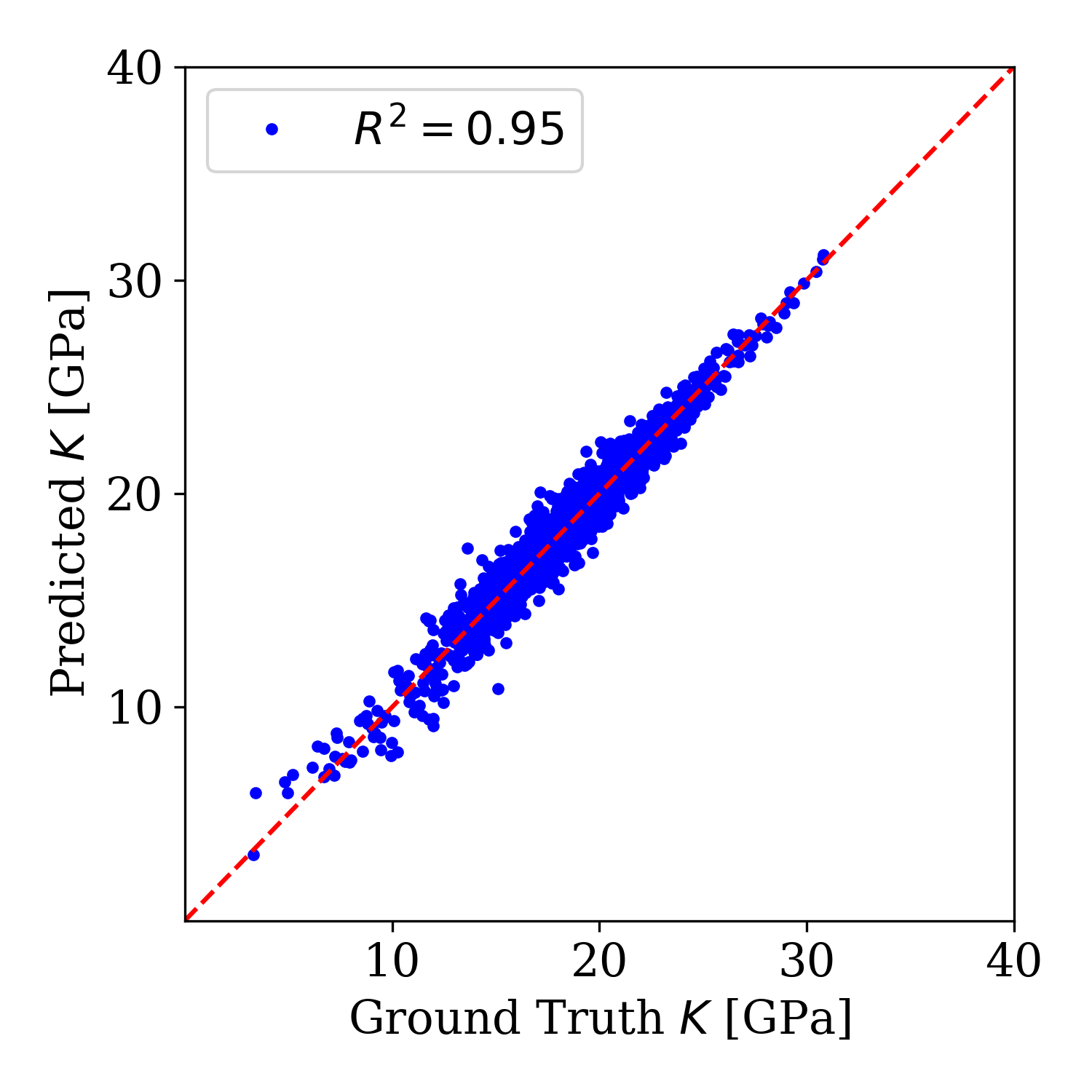}
        \caption{GNN bulk moduli prediction}
    \end{subfigure}
    \hfill
    \begin{subfigure}{0.45\textwidth}
        \centering\includegraphics[width=\textwidth]{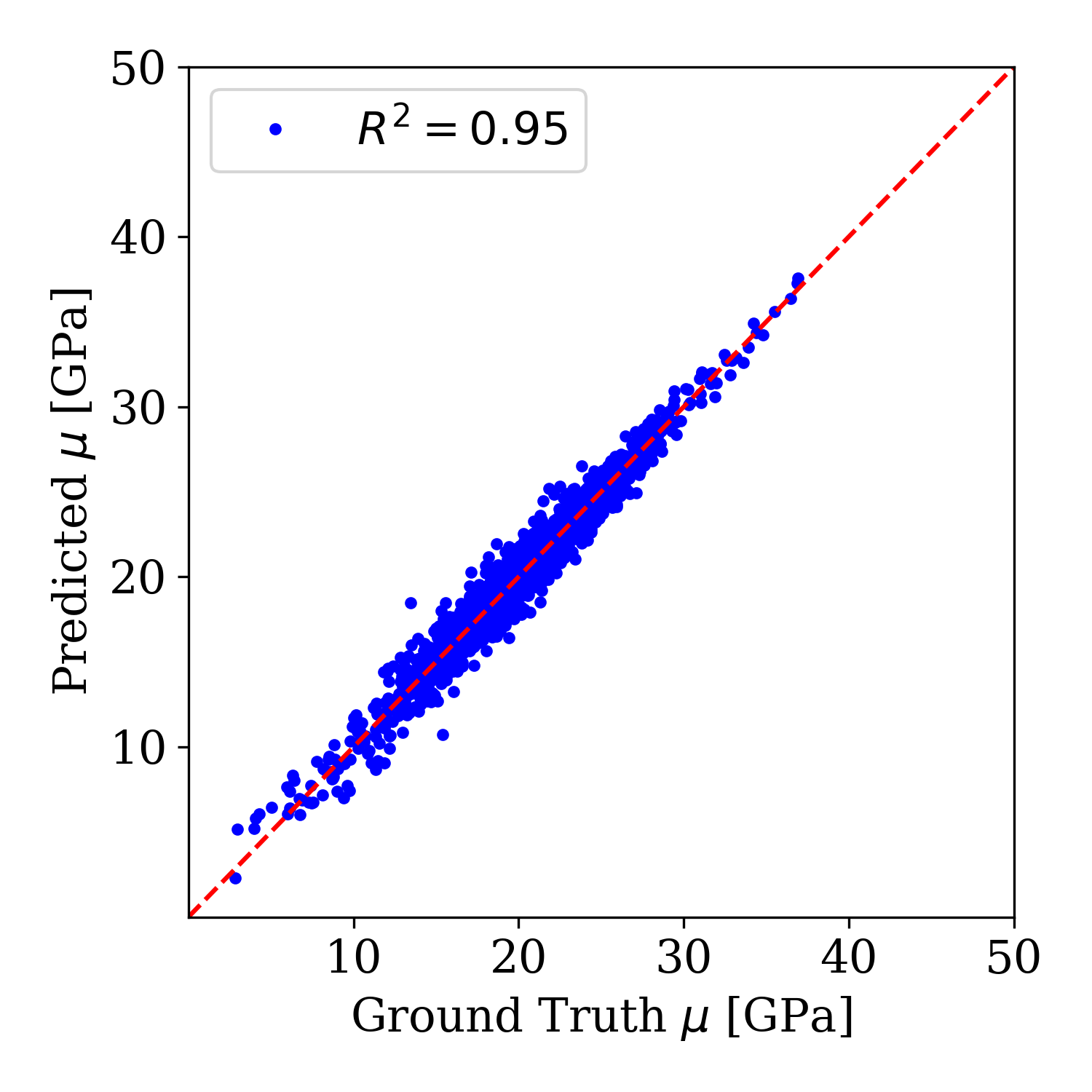}
        \caption{GNN shear moduli prediction}
    \end{subfigure}
    \vfill
    \begin{subfigure}{0.45\textwidth}
        \centering\includegraphics[width=\textwidth]{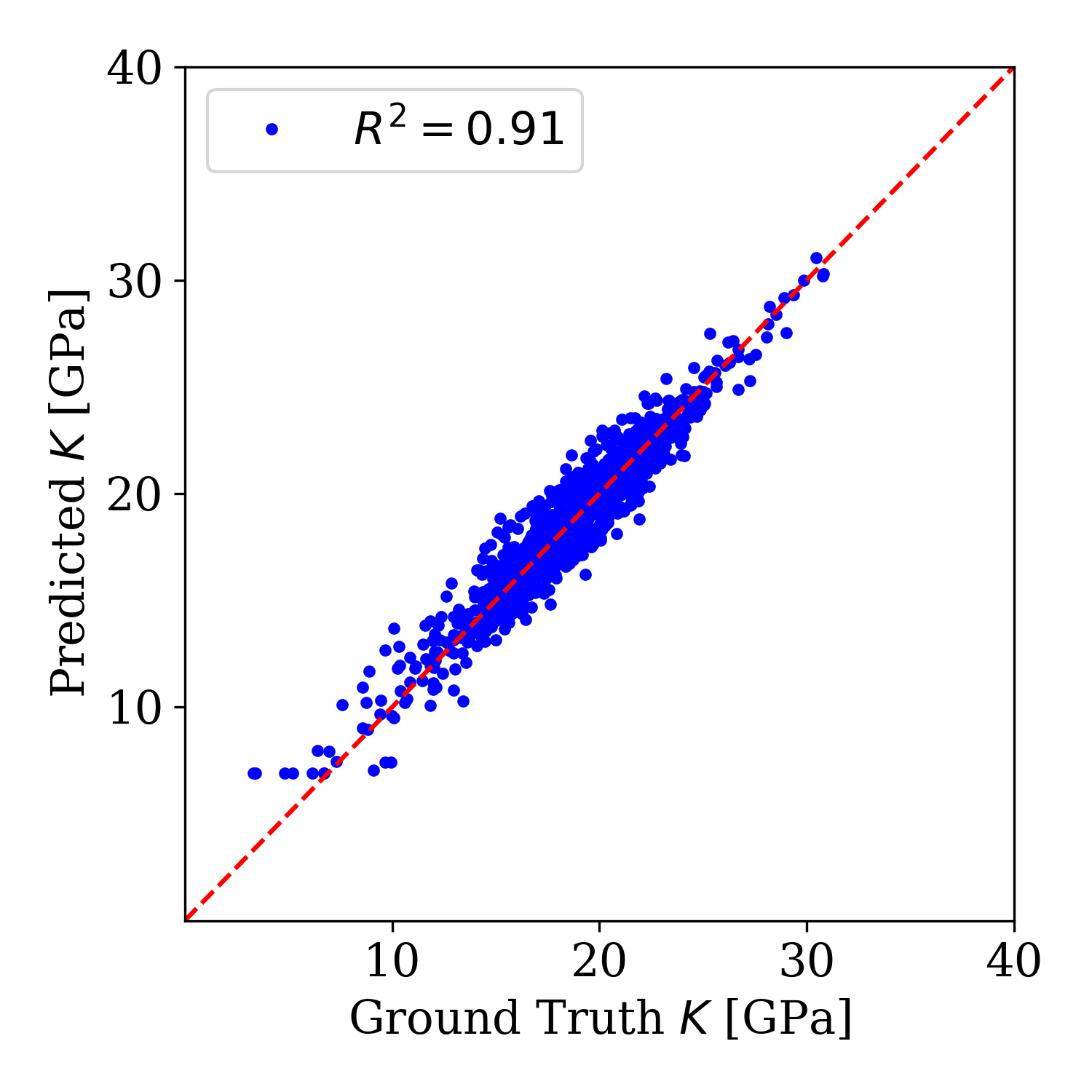}
        \caption{CNN bulk moduli prediction}
    \end{subfigure}
    \hfill
    \begin{subfigure}{0.45\textwidth}
        \centering\includegraphics[width=\textwidth]{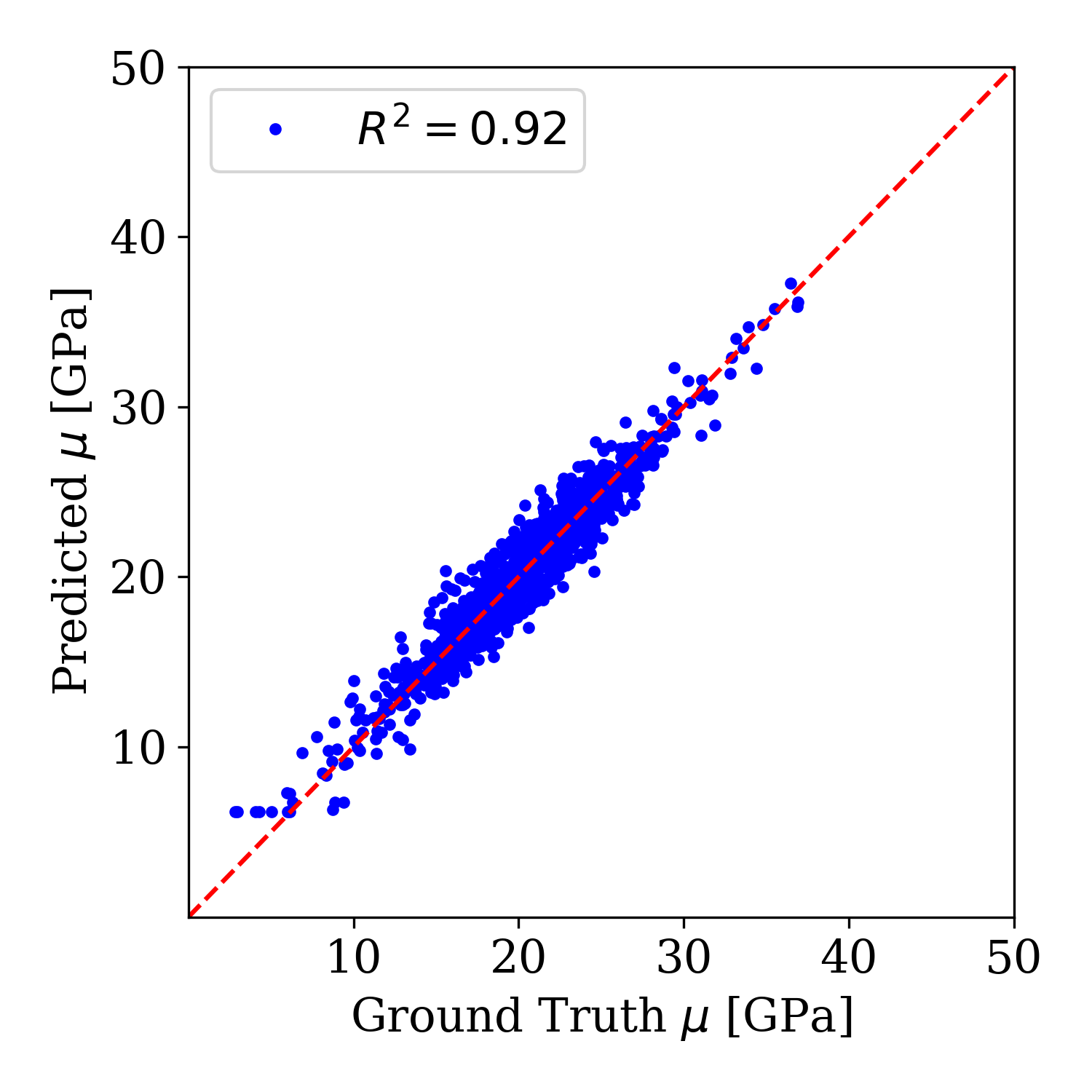}
        \caption{CNN shear moduli prediction}
    \end{subfigure}
    \caption{GNN and CNN prediction results for unseen rock (CG) for networks trained with 4 rocks (B1, B2, FB1, FB2)}
    \label{fig:GNN_vs_CNN_unseen_rock}
\end{figure}

Regarding prediction generality for unseen rocks, we assessed both the GNN and CNN models. In the previous section, the GNN, trained with three different image sizes, showed an ability to predict the properties of CG rocks with varying image sizes. However, to ensure these results derived from the GNN's expressiveness and not just a larger dataset, we trained both GNN and CNN models with four digital rocks (B1, B2, FB1, FB2) of single subcube image size (90). The comparison of these models reveals that the GNN achieved an $R^2$ value of 0.95 for $K$ and $\mu$. In contrast, the CNN predict the unseen properties with $R^2$ 0.91 and 0.92 for the same properties as shown in Figure \ref{fig:GNN_vs_CNN_unseen_rock}. This suggests that given the same amount of rock data, the GNN demonstrates superior capability in predicting the properties of unseen rocks. Combining the generalizability results from the previous section, GNN appears to capture underlying topological structures rather than just fitting from the images. 

\begin{figure}
\centering
\begin{subfigure}{0.45\textwidth}
\centering
\includegraphics[width=\textwidth]{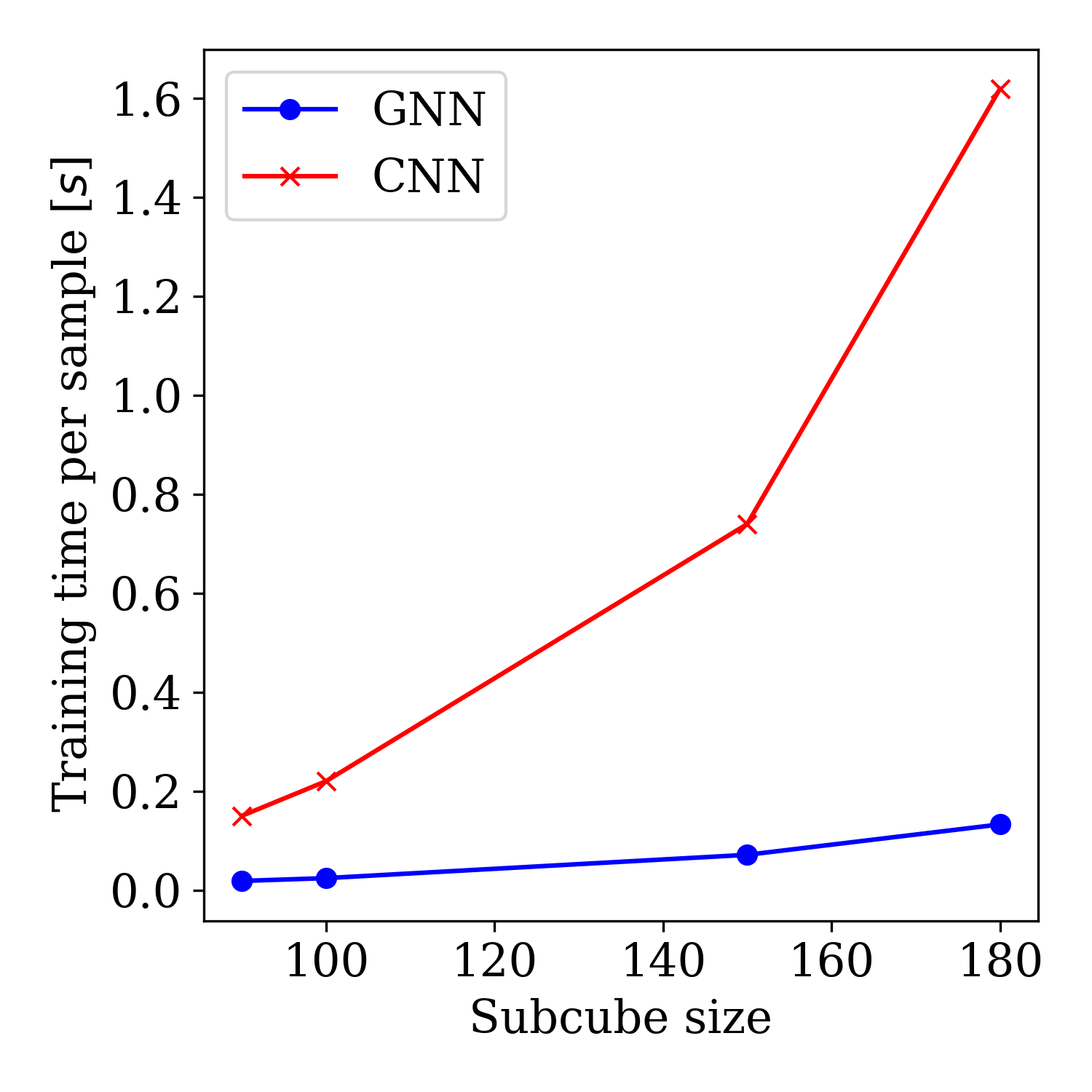}
\caption{Training time across different subcube sizes}
\label{fig:GNN_CNN_training_time}
\end{subfigure}
\hfill
\begin{subfigure}{0.45\textwidth}
\centering
\includegraphics[width=\textwidth]{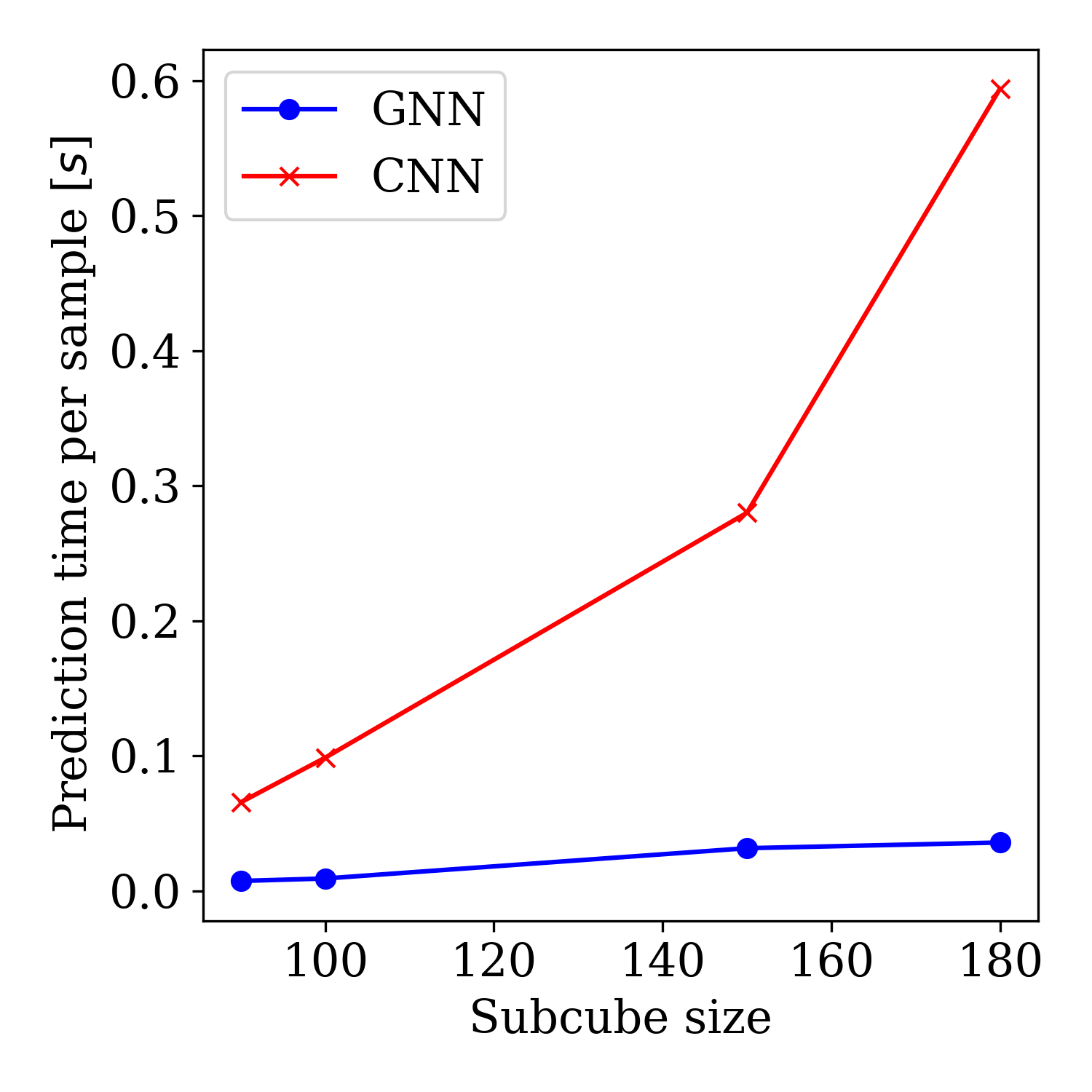}
\caption{Prediction time across different subcube sizes}
\label{fig:GNN_CNN_prediction_time}
\end{subfigure}
\caption{Comparison of GNN and CNN training and prediction time for a single sample across varying subcube sizes}
\label{fig:GNN_CNN_time}
\end{figure}

One of the key goals of employing deep learning methods in digital rock physics (DRP) is to replace time-consuming solution of the elasticity problems. Therefore, computational cost and time significantly influence the choice of the computational model. 
Although GPUs are used in our production runs, to ensure a fair comparison of the computational complexity of different approaches, here we perform benchmark tests on a single CPU.
We measure the time taken for training and prediction across varying subcube image sizes using a 32 GB memory CPU, as shown in Figure \ref{fig:GNN_CNN_time}. Based on the results, both the GNN and CNN models show increased training time with increasing subcube size (Figure \ref{fig:GNN_CNN_training_time}). For the GNN model, as the subcube size increases from 90 to 180, there's a 7.2 times increase in the training time per sample, from 0.018 to 0.133 seconds. In comparison, the CNN model experiences an even higher increase, with training time increasing by approximately 10.8 times, from 0.15 to 1.62 seconds. 
The increase of prediction times with increasing subcube size is more pronounced in CNN than in GNN.
Specifically, the CNN inference time increases by about 9.1 times when the subcube size grows from 90 to 180. Conversely, for the GNN, this increase is around 4.9 times. Thus, CNN becomes increasingly more expensive relative to GNN with the increase of the image size. Our analysis underscores GNN's advantage in terms of computational efficiency and scalability when dealing with large input data sizes, making it a promising tool for application in digital rock physics. 

Another essential factor in deep learning performance and efficiency is the flexibility of the batch size, which directly impacts the GPU memory consumption during model training. The batch size is a critical hyperparameter that governs both model performance and computational efficiency. Larger batch sizes can lead to improved training efficiency and potentially enhance the model's generalization performance by enabling exploration of a broader range of batch sizes \citep{masters2018revisiting}. We compared the GPU memory requirements for training GNN and CNN models with different batch sizes. Given the 80 GB GPU capacity of our test platform, GNNs demonstrated significantly better flexibility, accommodating a batch size of $2^{12}$, compared to CNNs, which were limited to a batch size of $2^{7}$ for a test subcube size of 90.

\begin{figure}
\centering
\includegraphics[width=0.45\textwidth]{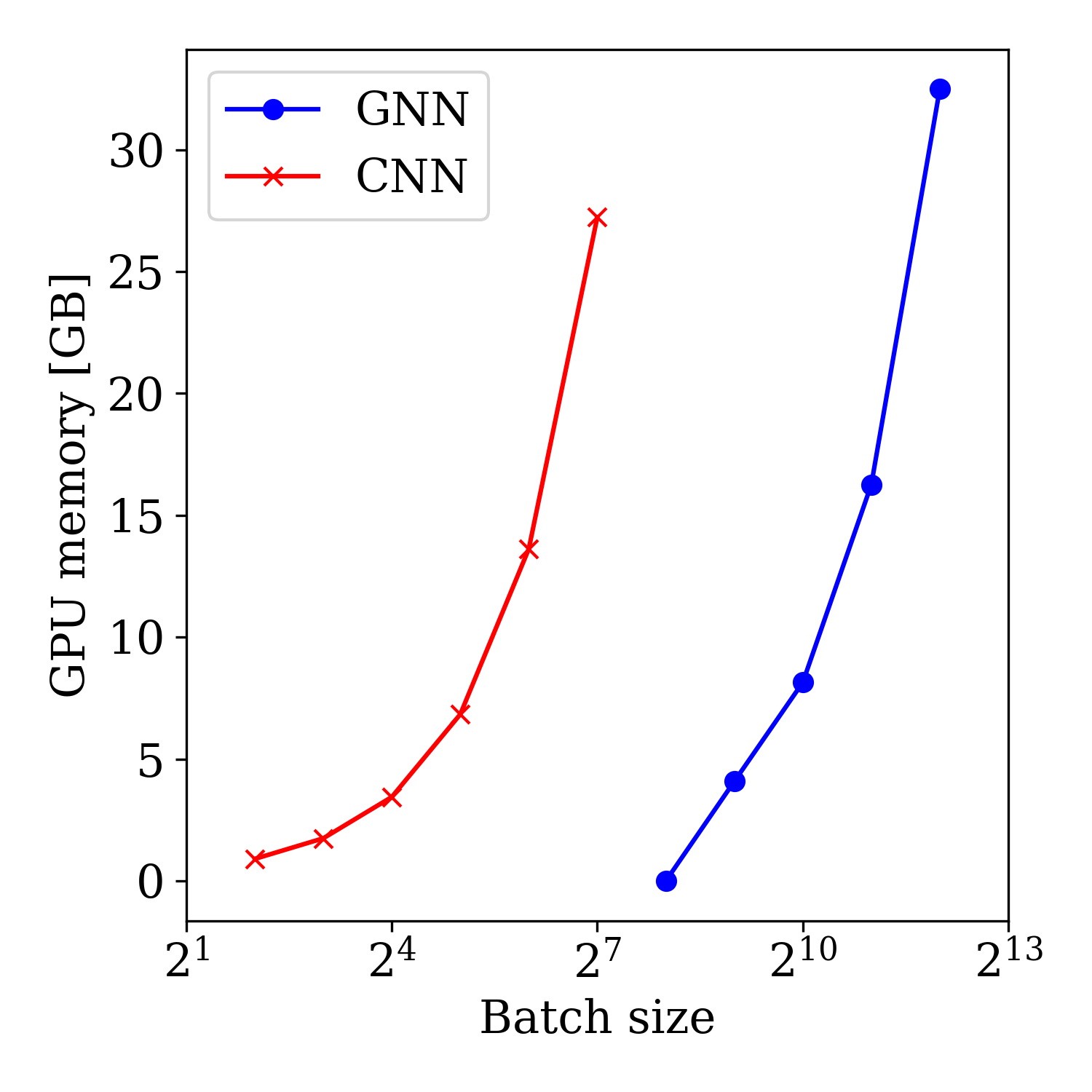}
\caption{GPU memory consumption for GNN and CNN models across different batch sizes}
\label{fig:GPU_batchsize_comparison}
\end{figure}

Figure \ref{fig:GPU_batchsize_comparison} shows the comparison between GPU memory consumption for different batch sizes in GNN and CNN models. The GNN model requires less than 0.1 GB of GPU memory for the batch size of $2^{7}$, while the CNN fails to train with that batch size on the same GPU. This ability to process larger batches results in higher training efficiency and provides more flexibility during the hyperparameter tuning phase. Furthermore, it allows GNNs to potentially enhance the model's performance by exploring a wider range of batch sizes.

In conclusion, our analysis reinforces the superiority of GNNs in terms of efficiency and generality given the digital rocks. While architectural modifications might enhance the CNN's performance, they invariably demand additional computational resources, presenting a practical limitation. This finding is in line with a growing body of research that highlights the significance of GNNs in processing complex, non-gridlike data structures in deep learning applications \cite{bronstein2017geometric}.

\subsection{Challenges of GNN preprocessing}
While our GNN-based approach has demonstrated its predictive capabilities for rock properties across various voxel sizes and unseen rock data, it's essential to address some of the inherent challenges associated with this application. One of the challenges is the preprocessing step required to transform voxel data into graph structures. Unlike CNNs, which can directly consume voxel or image data, GNNs necessitate this additional conversion step. In this work, we have used the Mapper algorithm for this transformation. 

\begin{figure}
\centering
\includegraphics[width=1\textwidth]{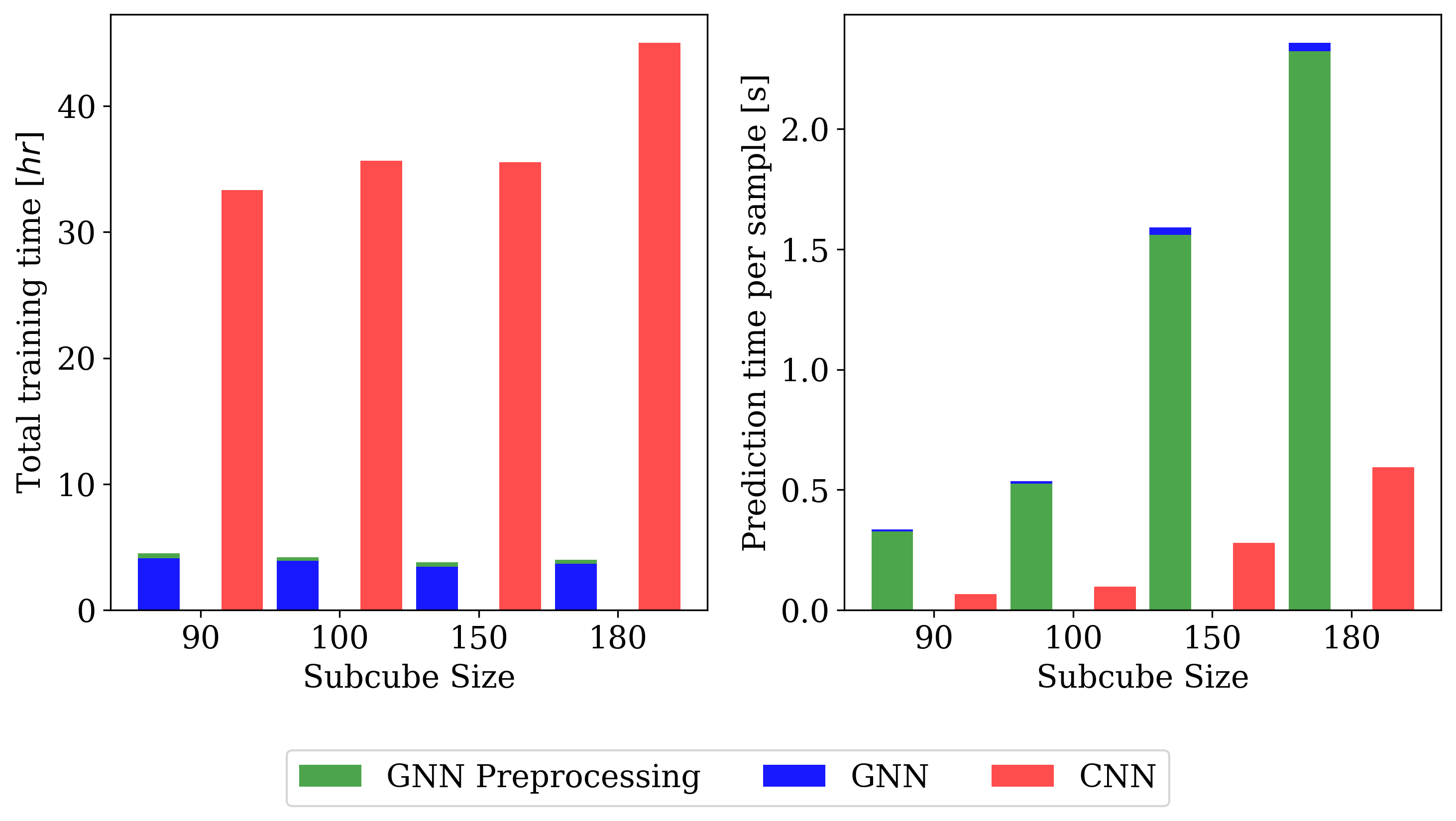}
\caption{Comparison of the GNN and CNN for total training duration and single-sample prediction time (The left figure represents results from training on four rock types (B1, B2, FB1, FB2) over 200 epochs.)}
\label{fig:Total_training_time}
\end{figure}

For instance, generating a $1000$ graph dataset with a subcube size of $90$ from a $900^3$ voxel requires $295.1$ seconds using a single CPU (most conservative case). Figure~\ref{fig:Total_training_time} shows the total training times of GNN and CNN over 200 epochs across the subcube sizes. While GNN's training time, including the preprocessing, remains 7.4 and 11.2 times shorter for the subcube size from 90 to 180 compared to the CNN's training time, it's notable that when pre-trained CNN models are available for each voxel size, the prediction time for GNN is longer than CNN due to the need to convert the voxel data to a graph.  In particular, prediction with GNN takes 5.1 times more than the CNN for a subcube size of 90 and 3.9 times for a subcube size of 180. Thus, there is still room for improvement to exploit the advantages of GNN, such as multi-CPU parallelization and GPU-based accelerated graph construction. In our future study, we will focus on accelerating the data transformation process to improve the efficiency of GNNs, thus mitigating the preprocessing challenge.

\section{Conclusion}
\label{section:Conclusion}
In this research, we proposed a novel GNN methodology to predict the elastic moduli of digital rocks. We constructed a graph dataset using the Mapper algorithm, providing a simplified yet valuable representation of microstructures. We then assessed the mapper-constructed graphs by the random forest regressor to determine whether it could predict elastic moduli, even in the absence of node features.

Our application of GNNs showed considerable promise in predicting elastic moduli. The results indicated that GNNs could handle different graph sizes, which were created from various subcube sizes. In addition, the GNN model demonstrated strong performance not only on the testing dataset but also exhibited a high degree of generalization. It successfully predicted the elastic moduli of unseen rocks and subcube sizes.

In comparison with CNN, the GNN method showed superior performance in predicting the elastic moduli of unseen rocks. Furthermore, the GNN method demonstrated a moderate increase in training and prediction times as the subcube size grew, in contrast to the exponential increase observed with CNNs, thereby indicating superior scalability. Another significant advantage of GNNs was observed in their capacity to train with larger batch sizes. This flexibility in model training is a crucial aspect of deep learning methodologies. In comparison, CNNs faced limitations in their batch size due to GPU memory constraints. This capacity of GNNs underscores their potential as efficient tools for handling large-scale data. However, the remaining challenge is the GNN's need to transform voxel data into graph structures. While GNNs excel over CNNs concerning overall training duration, their prediction times lag behind those of CNNs. Future studies focusing on optimizing this conversion process can potentially enhance the efficiency of GNNs even further.
In summary, this research highlights the utility of GNNs as an effective tool in predicting rock properties and elucidating the complex relationships between pore geometries and properties.

\section*{Data Availability Statement}
The code and data used in this study are freely available and can be accessed on the GitHub repository at \url{https://github.com/jh-chung1/GNN_ElasticModulus_Prediction}.

\section*{Acknowledgments}
We acknowledge Shell for financial support and for providing the digital rock images. This study has mostly been performed using the Sherlock cluster at Stanford University. We are grateful to Stanford University and the Stanford Research Computing Center for providing computational resources and support in this research. We acknowledge the sponsors of the Stanford Center for Earth Resources Forecasting (SCERF). The authors also would like to thank Math2Market for providing the GeoDict software at a discount and technical support.

\newpage
\appendix

\section{Graph Topological Metrics}
\label{section:Appendix_graph_topo_metrics}
\subsection{Vertex degree}
Given an undirected graph \(G(V, E)\), the set of neighbors of a vertex \(v\) is denoted as \(\mathcal{N}(v)\). The degree of a vertex \(deg(v)\) is defined as the number of neighbors that \(v\) has, as in equation \ref{eqn:vertex_degree}. The average vertex degree of the graph \(deg(G)\) is the average number of neighbors over all vertices (Equation \ref{eqn:graph_vertex_degree}).

\begin{equation}
\label{eqn:vertex_degree}
deg(v) = |\mathcal{N}(v)|
\end{equation}

\begin{equation}
\label{eqn:graph_vertex_degree}
deg(G) := \frac{1}{|V|}\Sigma_{v \in V} deg(v)
\end{equation}

\subsection{Closeness Centrality}
Closeness Centrality of a vertex \(v\) measures how close the vertex is to all other vertices in the graph, defined as the inverse of the sum of the shortest path distances from \(v\) to all other vertices. Given a graph \(G(V, E)\), the closeness centrality \(C_C(v)\) of a vertex \(v\) can be computed as in equation \ref{eqn:closeness}.

\begin{equation}
\label{eqn:closeness}
C_C(v) := \frac{1}{\sum_{u \in V, u \neq v} d(u, v)}
\end{equation}

where \(d(u, v)\) is the shortest path distance between vertices \(u\) and \(v\).

\subsection{Eigenvector Centrality}
Eigenvector centrality assesses the influence of a vertex within a network based on the notion that connections to influential vertices confer greater importance \citep{bonacich2007some}. For a graph \(G(V, E)\), the computation involves an iterative process where each vertex's centrality is updated based on the centrality of its neighbors, and then normalized. The final values correspond to the eigenvector associated with the largest eigenvalue of the adjacency matrix of the graph. Mathematically, the iterative relation is given by:
\begin{equation}
\label{eqn:eigenvector}
EC(v) := \frac{1}{\lambda} \Sigma_{u \in \mathcal{N}(v)} EC(u)
\end{equation}
where \(\lambda\) is a constant, often the largest eigenvalue of the adjacency matrix, and \(\mathcal{N}(v)\) denotes the neighbors of vertex \(v\).

\subsection{Pagerank}
Pagerank, initially devised for web search ranking, determines the importance of vertices in a network \citep{langville2004deeper}. It models the likelihood of a random walker visiting a particular vertex. For a graph \(G(V, E)\), the computation involves iteratively updating each vertex's rank based on its inbound links, followed by normalization. The recursive relation for each iteration is:
\begin{equation}
\label{eqn:pagerank}
PR(v) = (1 - d) + d \times \Sigma_{u \in \mathcal{N}(v)} \frac{PR(u)}{deg(u)}
\end{equation}
where \(d\) is a damping factor, \(\mathcal{N}(v)\) signifies the vertices linking to \(v\), and \(deg(u)\) is the out-degree of vertex \(u\).

\section{Traditional effective elastic modulus computation}
\label{section:Appendix_DEM_results}
\subsection{DEM prediction performance on test dataset}
We utilize Differential Effective Medium (DEM) models \citep{berryman1980long, norris1985effective} to compute the effective properties and compare the performance of graph-based predictions. The DEM model computes the properties of a two-phase composite by starting with an initial background matrix (phase 1) and incrementally adding inclusions of another phase (phase 2). For our purposes, phase 1 is quartz mineral and phase 2 consists of pores filled with air. The system of ordinary differential equations governing the effective moduli is as follows \citep{mavko2020rock}:

\begin{equation}
\begin{aligned}
(1-y)\frac{d}{dy}[K^*(y)]&=(K_{2} - K^*)P^{*2}(y)\\
(1-y)\frac{d}{dy}[\mu^*(y)]&=(\mu_{2} - \mu^*)Q^{*2}(y)\\
P^{mi} &= \frac{K_m + \frac{4}{3}\mu_i}{K_i + \frac{4}{3}\mu_i + \pi \alpha \beta_m}\\
Q^{mi} &= \frac{1}{5}\left[1 + \frac{8\mu_m}{4\mu_i + \pi \alpha \left(\mu_m+2\beta_m\right)}+2\frac{K_i + \frac{2}{3}\left(\mu_i+\mu_m\right)}{K_i+\frac{4}{3}\mu_i+\pi\alpha\beta_m}\right]
\end{aligned}
\end{equation}

where the $K^*$ and $\mu^*$ are effective bulk and shear moduli, respectively, and $P^{*2}$ and $Q^{*2}$ are geometric factors for the pore inclusion with the background medium with effective moduli $K^*$ and $\mu^*$. $y$ is porosity. The subscript $m$ and $i$ denote the background and inclusion material. In this study, we assumed penny cracks for the pore inclusion shapes. For more details, please refer to \citep{berryman1995mixture}.

\begin{figure}[pos=h!]
\centering
\includegraphics[width=0.45\textwidth]{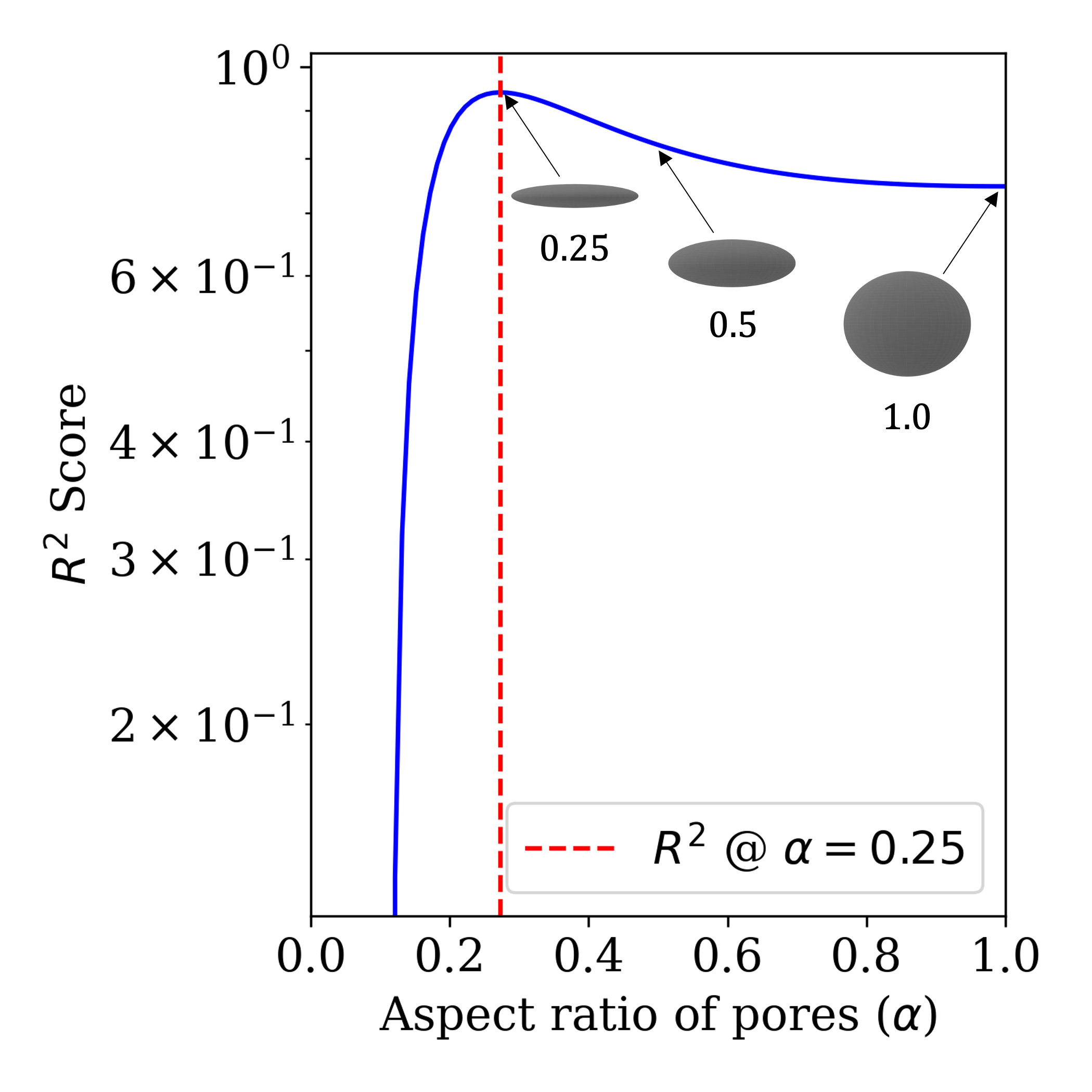}
\caption{DEM prediction performance with aspect ratio variation}
\label{fig:DEM_optimization}
\end{figure}

\begin{figure}[pos=h!]
    \centering
    \begin{subfigure}{0.45\textwidth}
    \centering
        \includegraphics[width=\textwidth]{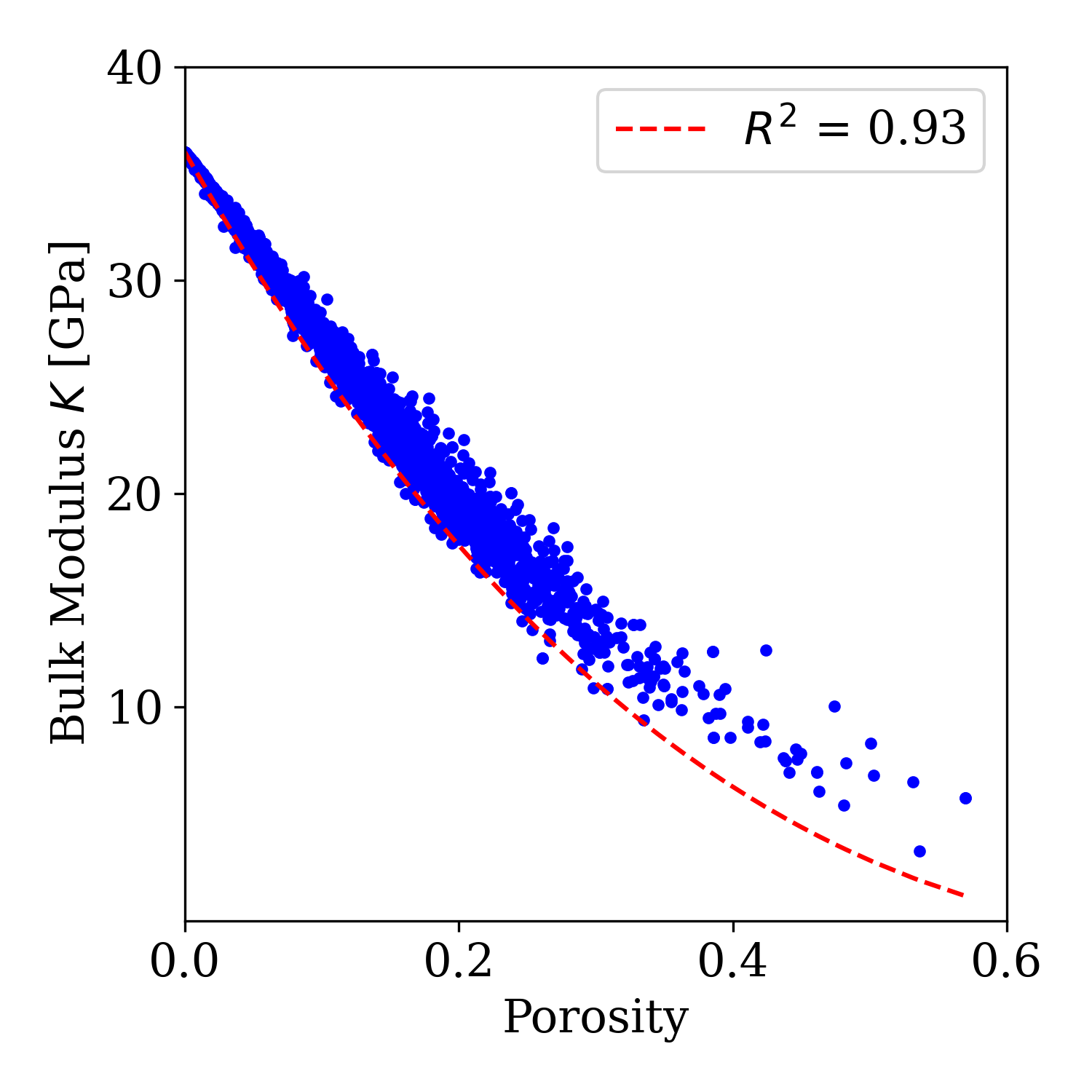}
        \caption{Bulk moduli prediction}
    \end{subfigure}
    \hspace{0.1cm}
    \begin{subfigure}{0.45\textwidth}
    \centering
        \includegraphics[width=\textwidth]{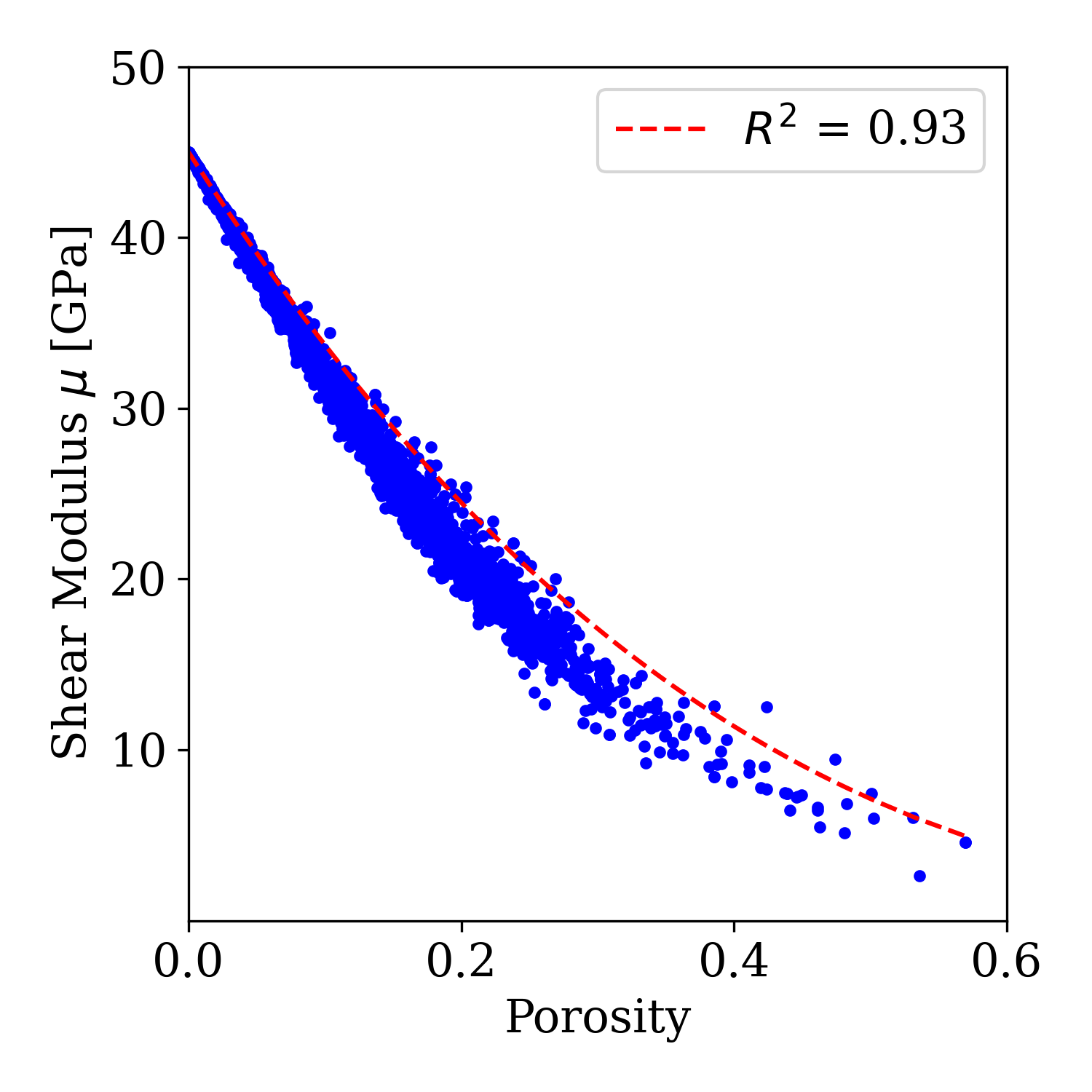}
        \caption{Shear moduli prediction}
    \end{subfigure}
    \caption{DEM prediction results with porosity variation}
    \label{fig:DEM_test_results}
\end{figure}

To ensure a fair comparison, we optimized the aspect ratio ($\alpha$) to maximize the average of the $R^2$ scores for bulk and shear moduli, using the same training dataset as the random forest model described in Section \ref{section:random_forest}. Figure \ref{fig:DEM_optimization} shows the variation of the averaged $R^2$ with different pore aspect ratios, indicating maximum $R^2$ when the aspect ratio is 0.25. Figure \ref{fig:DEM_test_results} shows the optimized DEM prediction results with porosities for the testing dataset.

\subsection{Hashin-Shtrikman bounds}
We employ the Hashin-Shtrikman bounds \citep{hashin1963variational} for evaluating qualitative performances of the graph-based predictions. The Hashin-Shtrikman bounds provide the narrowest bounds on the isotropic effective moduli without specifying geometrical information. The bounds for mineral and pore constituents are as follows \citep{mavko2020rock}:

\begin{equation}
\begin{aligned}
K^{HS+} &= K_{mineral} + \frac{\phi}{(K_{pore}-K_{mineral})^{-1}+(1-\phi)\,(K_{mineral}+\frac{4}{3}\mu_{mineral})^{-1}}\\
K^{HS-} &= K_{pore} + \frac{1-\phi}{(K_{mineral}-K_{pore})^{-1}+\phi\,(K_{pore}+\frac{4}{3}\mu_{pore})^{-1}}\\
\mu^{HS+} &= \mu_{mineral} + \frac{\phi}{(\mu_{pore}-\mu_{mineral})^{-1}+2\,(1-\phi)\,(K_{mineral}+2\mu_{mineral})\,\left[5\mu_{mineral}\left(K_{mineral}+\frac{4}{3}\mu_{mineral}\right)\right]^{-1}}\\
\mu^{HS-} &= \mu_{pore} + \frac{1-\phi}{(\mu_{mineral}-\mu_{pore})^{-1}+2\,\phi\,(K_{pore}+2\mu_{pore})\,\left[5\mu_{pore}\left(K_{pore}+\frac{4}{3}\mu_{pore}\right)\right]^{-1}}
\end{aligned}
\end{equation}
where the subscripts $HS^+$ and $HS^-$ denote the upper and lower bounds, respectively.

\section{CNN prediction performance on test dataset}
\label{section:Appendix_cnn_results}
\begin{figure}
    \centering
    \begin{subfigure}{0.45\textwidth}
    \centering
        \includegraphics[width=\textwidth]{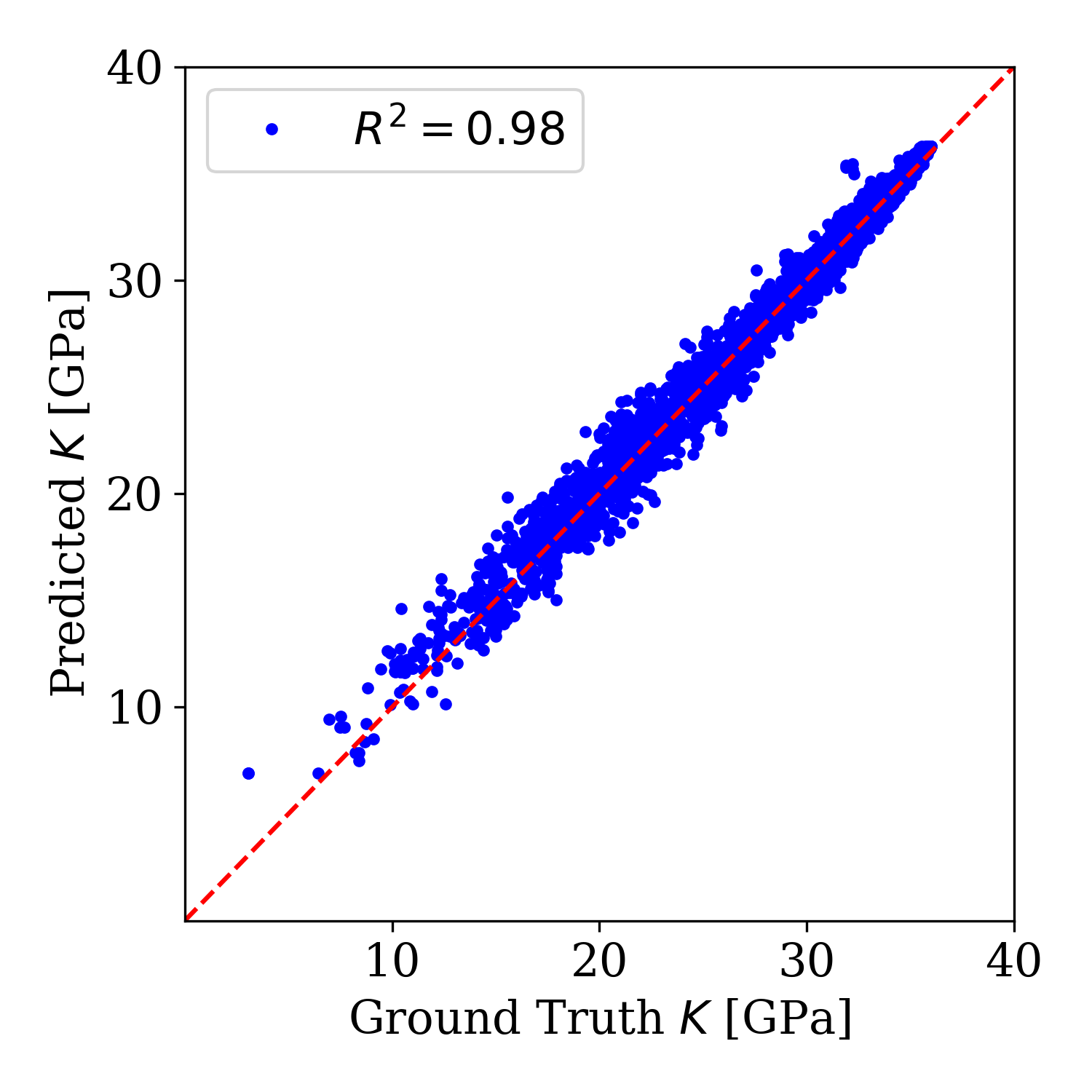}
        \caption{Bulk moduli prediction}
    \end{subfigure}
    \hspace{0.1cm}
    \begin{subfigure}{0.45\textwidth}
    \centering
        \includegraphics[width=\textwidth]{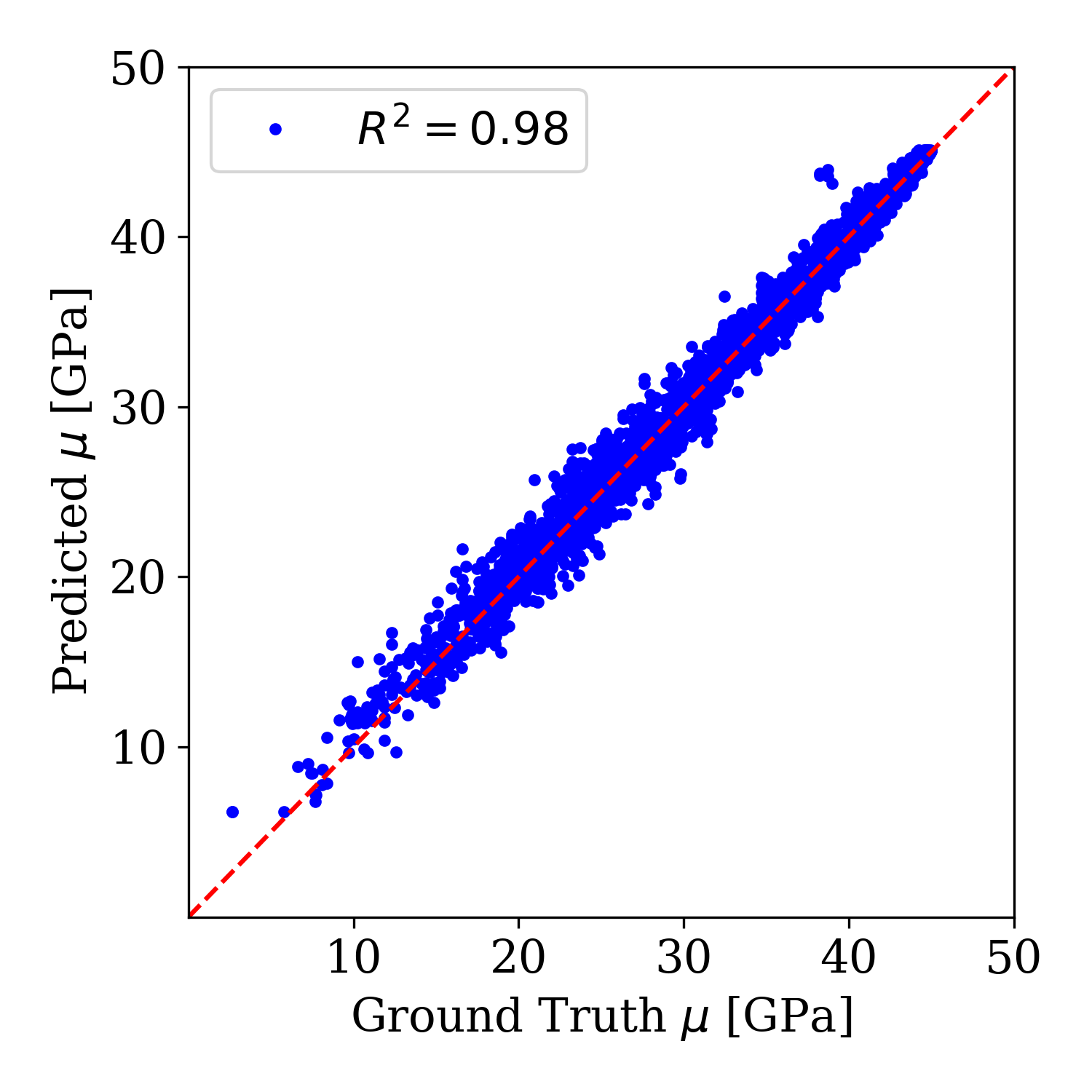}
        \caption{Shear moduli prediction}
    \end{subfigure}
    \caption{CNN testing results for trained with 4 rocks (B1, B2, FB1, FB2) and subcube size 90}
    \label{fig:CNN_test_results}
\end{figure}

\newpage

\bibliographystyle{cas-model2-names}
\bibliography{bibliography} 

\end{document}